 \documentclass[pmlr,twocolumn]{jmlr} 

 \usepackage{bbm}

\usepackage{booktabs}
\usepackage[load-configurations=version-1]{siunitx} 

\theorembodyfont{\upshape}
\theoremheaderfont{\scshape}
\theorempostheader{:}
\theoremsep{\newline}

\jmlrvolume{LEAVE UNSET}
\jmlryear{2020}
\jmlrsubmitted{LEAVE UNSET}
\jmlrpublished{LEAVE UNSET}
\jmlrworkshop{Machine Learning for Health (ML4H) 2020} 

\RequirePackage{import}
\usepackage{amsmath}
\usepackage{amssymb}
\usepackage{bm}

\newcommand{\mathbold}[1]{\ensuremath{\boldsymbol{\mathbf{#1}}}}

\newcommand{\mbalpha}{\mathbold{\alpha}}
\newcommand{\mbbeta}{\mathbold{\beta}}

\newcommand{\mbmu}{\mathbold{\mu}}

\newcommand{\mbsigma}{\mathbold{\sigma}}

\newcommand{\mbtheta}{\mathbold{\theta}}

\newcommand{\bma}{\bm{a}}
\newcommand{\bmb}{\bm{b}}

\newcommand{\bmh}{\bm{h}}

\newcommand{\bmk}{\bm{k}}

\newcommand{\bmn}{\bm{n}}

\newcommand{\bmq}{\bm{q}}
\newcommand{\bmr}{\bm{r}}
\newcommand{\bms}{\bm{s}}

\newcommand{\bmv}{\bm{v}}
\newcommand{\bmw}{\bm{w}}
\newcommand{\bmx}{\bm{x}}

\newcommand{\bmz}{\bm{z}}

\newcommand{\bmR}{\bm{R}}

\newcommand{\bmV}{\bm{V}}
\newcommand{\bmW}{\bm{W}}

\newcommand{\mcH}{\mathcal{H}}

\newcommand{\mcM}{\mathcal{M}}

\newcommand{\mcQ}{\mathcal{Q}}

\newcommand{\mcV}{\mathcal{V}}
\newcommand{\mcW}{\mathcal{W}}
\newcommand{\mcX}{\mathcal{X}}

\newcommand{\mcZ}{\mathcal{Z}}

\newcommand{\mbbE}{\mathbb{E}}

\newcommand{\mbbR}{\mathbb{R}}

\DeclareMathOperator*{\softmax}{softmax}

\DeclareMathOperator*{\relu}{ReLU}

\DeclareMathOperator*{\sech}{sech}
\usepackage[acronym,smallcaps,nowarn,section,nonumberlist]{glossaries}
\glsdisablehyper
\makeglossaries

\setacronymstyle{long-short}

\newacronym{argat}{ARGAT}{Across-Relation Graph Attention}
\newacronym{auc}{AUC}{Area Under the Curve}

\newacronym{bert}{BERT}{Bidirectional Encoder Representations from Transformers}
\newacronym{bow}{BOW}{Bag of Words}
\newacronym{boe}{BOE}{Bag of Embeddings}
\newacronym{borep}{BOREP}{Bag of Random Embedding Projections}
\newacronym{bn}{BN}{Bayesian Network}

\newacronym{cdf}{CDF}{Cumulative Distribution Function}
\newacronym{cnn}{CNN}{Convolutional Neural Network}
\newacronym{cvae}{CVAE}{Conditional Variational Autoencoder}
\newacronym{dag}{DAG}{Directed Acyclic Graph}
\newacronym{dgm}{DGM}{Directed Graphical Model}
\newacronym{dnn}{DNN}{Deep Neural Network}
\newacronym{ecfp4}{ECFP4}{Extended Connectivity Fingerprint}
\newacronym{ehr}{EHR}{Electronic Health Record}
\newacronym{esn}{ESN}{Echo State Network}

\newacronym{gat}{GAT}{Graph Attention Network}
\newacronym{gcn}{GCN}{Graph Convolutional Network}
\newacronym{gdl}{GDL}{Geometric Deep Learning}
\newacronym{ggnn}{GGNN}{Gated Graph Neural Network}
\newacronym{glove}{GloVe}{Global Vectors for Word Representation}
\newacronym{gnn}{GNN}{Graph Neural Network}
\newacronym{gru}{GRU}{Gated Recurrent Unit}

\newacronym{hmm}{HMM}{Hidden Markov Model}
\newacronym{kld}{KLD}{Kullback-Leibler Divergence}
\newacronym{knn}{KNN}{K-Nearest Neighbour}
\newacronym{idf}{IDF}{Inverse Document Frequency}
\newacronym{ig}{IG}{Information Gain}
\newacronym{isan}{ISAN}{Input Switched Affine Network}
\newacronym{lhs}{L.H.S.}{left hand side}
\newacronym{lstm}{LSTM}{Long Short Term Memory Network}

\newacronym{map}{MAP}{Maximum a Posteriori}
\newacronym{mc}{MC}{Monte Carlo}
\newacronym{mdl}{MDL}{Minimum Description Length}
\newacronym{ml}{ML}{Machine Learning}
\newacronym{mlp}{MLP}{Multi-Layer Perceptron}
\newacronym{mle}{MLE}{Maximum Likelihood Estimation}
\newacronym{mnist}{MNIST}{Modified National Institute of Standards and Technology}

\newacronym{nlp}{NLP}{Natural Language Processing}
\newacronym{nn}{NN}{Neural Network}
\newacronym{nll}{NLL}{Negative Log Likelihood}
\newacronym{ode}{ODE}{Ordinary Differential Equation}
\newacronym{oh}{OH}{One-Hot}

\newacronym{pdf}{PDF}{Probability Density Function}
\newacronym{pgm}{PGM}{Probabilistic Graphical Model}

\newacronym{relu}{ReLU}{Rectified Linear Unit}
\newacronym{rdf}{RDF}{Resource Description Framework}
\newacronym{rgb}{RGB}{Red Green Blue}
\newacronym{rgc}{RGC}{Relational Graph Convolution}
\newacronym{rgat}{RGAT}{Relational Graph Attention Network}
\newacronym{rgcn}{RGCN}{Relational Graph Convolutional Network}
\newacronym{rhs}{R.H.S.}{Right Hand Side}
\newacronym{rnn}{RNN}{Recurrent Neural Network}
\newacronym{roc}{ROC}{Receiver Operating Characteristic}

\newacronym{sgd}{SGD}{Stochastic Gradient Descent}
\newacronym{sg}{SG}{SkipGram}
\newacronym{st}{ST}{SkipThought}
\newacronym{sts}{STS}{Semantic Textual Similarity}

\newacronym{termfreq}{TF}{Term Frequency}
\newacronym{tfidf}{TF-IDF}{Term Frequency-Inverse Document Frequency}
\newacronym[plural=TPPs, descriptionplural={Temporal Point Processes}]{tpp}{TPP}{Temporal Point Process}

\newacronym{vae}{VAE}{Variational Autoencoder}

\newacronym{wirgat}{WIRGAT}{Within-Relation Graph Attention}
\newacronym{wl}{WL}{Weisfeiler-Lehman}

\newcommand{\dhid}{D_{\textrm{Hid}}}
\newcommand{\demb}{D_{\textrm{Emb}}}

\usepackage{tikz}
\usetikzlibrary{arrows,positioning} 
\tikzset{
    >=stealth',
    punkt/.style={
           rectangle,
           rounded corners,
           draw=black, very thick,
           minimum width=4.5em,
           minimum height=2em,
           text centered},
    pil/.style={
           ->,
           thick,
           shorten <=2pt,
           shorten >=2pt,}
}
\tikzset{>=latex}

\usepackage{multicol}
\usepackage{cleveref}
\newcommand{\Lim}[1]{\raisebox{0.5ex}{\scalebox{0.8}{$\displaystyle \lim_{#1}\;$}}}

\title[Neural TPPs for Modelling Electronic Health Records]{Neural Temporal Point Processes \titlebreak For Modelling Electronic Health Records}

\author{%
\Name{Joseph Enguehard}\thanks{Equal contribution.
  Joseph Enguehard and Dan Busbridge led the research, implemented the models and performed the experiments.
  Adam Bozson created the synthetic EHR datasets and provided insights during project inception.
  Claire Woodcock strategically guided the project and contributed the broader impact statement.
  Nils Y. Hammerla provided technical guidance throughout.} \Email{joseph.enguehard@babylonhealth.com}\\
\addr Babylon Health
\AND
\Name{Dan Busbridge}$^*$\\
\addr Babylon Health
\AND
\Name{Adam Bozson}\\
\addr Babylon Health
\AND
\Name{Claire Woodcock}\\
\addr Babylon Health
\AND
\Name{Nils Hammerla}\\
\addr Babylon Health
}

\begin{document}
\maketitle
\begin{abstract}
  The modelling of Electronic Health Records (EHRs) has the potential to drive more efficient allocation of healthcare resources, enabling early intervention strategies and advancing personalised healthcare. 
  However, EHRs are challenging to model due to their realisation as noisy, multi-modal data occurring at irregular time intervals. 
  To address their temporal nature, we treat EHRs as samples generated by a Temporal Point Process (TPP), enabling us to model what happened in an event with when it happened in a principled way. 
  We gather and propose neural network parameterisations of TPPs, collectively referred to as Neural TPPs. 
  We perform evaluations on synthetic EHRs as well as on a set of established benchmarks. 
  We show that TPPs significantly outperform their non-TPP counterparts on EHRs. 
  We also show that an assumption of many Neural TPPs, that the class distribution is conditionally independent of time, reduces performance on EHRs. 
  Finally, our proposed attention-based Neural TPP performs favourably compared to existing models, whilst aligning with real world interpretability requirements, an important step towards a component of clinical decision support systems.
\end{abstract}

\section{Introduction}

Healthcare systems today are under intense pressure. Costs of care are increasing, resources are constrained, and outcomes are worsening \citep{topolHighperformanceMedicineConvergence2019}. 
Given these intense challenges, healthcare stands as one of the most promising applications of \gls{ml}.
In particular, interest in modelling \glspl{ehr} has recently increased \citep{islamMarkedPointProcess2017,shickelDeepEHRSurvey2018,darabiUnsupervisedRepresentationEHR2019,liBEHRTTransformerElectronic2019,rodrigues-jrPatientTrajectoryPrediction2019}. 
Better EHR modelling at an institutional level could enable more effective allocation of healthcare \citep{gijsbertsRaceEthnicDifferences2015}. 
At a clinical level accurate models could provision tools which improve patient outcomes by automating labour intensive administrative tasks \citep{Dhruva2020} and developing early and optimal intervention strategies for doctors \citep{Komorowski2019}. 
However, facets of health evolve at differing rates, and \glspl{ehr} are typically realised as noisy, multi-modal data occurring at irregular time intervals. 
This makes \glspl{ehr} difficult to model using common \gls{ml} methods. 

We propose to address the temporal nature of longitudinal \glspl{ehr}\footnote{Longitudinal \glspl{ehr} consist of comprehensive records of patients' clinical experience, usually spanning over several decades. This type of data does not include ICU datasets, such as MIMIC-III \citep{mimiciii} which focus on a shorter time frame.} by treating them as samples generated by a \gls{tpp}, a probabilistic framework which can deal with data occurring at irregular time intervals.
Specifically, we use a \gls{nn} to approximate its density, which specifies the probability of the next event happening at a given time.
We call these models Neural \glspl{tpp}. 
We propose jointly modelling times and labels to capture the varying evolution rates within \glspl{ehr}.

To test this, aware of the need for transparency in research \citep{pineauImprovingReproducibilityMachine2020} and for sensitive handling of health data \citep{kalkmanResponsibleDataSharing2019}, we perform an extensive study on synthetic \gls{ehr} datasets generated using the open source Synthea simulator \citep{walonoskiSyntheaApproachMethod2018}. 
For completeness, we also perform evaluations on benchmark datasets commonly used in the \gls{tpp} literature. We find that:
\begin{itemize}
    \item Our proposed Neural \glspl{tpp}, where labels are jointly modelled with time, significantly outperform those that treat them as conditionally independent; a common simplification in the \gls{tpp} literature,
    \item Particularly in the case of \glspl{tpp} on labelled data, metrics that decompose the performance in terms of both label and temporal accuracy should be reported,
    \item Some datasets used for benchmarking \glspl{tpp} are easily solved by a time-independent model. On other datasets, including synthetic \glspl{ehr}, \glspl{tpp} outperform their non-\gls{tpp} counterparts.
\end{itemize}
Finally, we present a Neural \gls{tpp}  whose attention-based mechanism provides interpretable information without compromising on performance. This is an essential contribution in the development of research for clinical applications. 
The wider impact of pursuing this direction is detailed in \Cref{sec:impact}.

\section{Background}
\subsection{Temporal point processes}
A \gls{tpp} is a random process that generates a sequence of $N$ events
$\mcH=\{(t_i,\mcM_i)\}_{i=1}^N$
within a given observation window
$t_i\in[w_-,w_+]$.
Each event consists of a set of labels
$\mcM_i\subseteq\{1,\ldots,M\}$
localised at times
$t_{i-1}<t_{i}$.
Labels may be independent or mutually exclusive, depending on the task.

A \gls{tpp} is fully characterised through its conditional intensity $\lambda_m^*(t)$, for $t_{i-1}<t\leq t_{i}$:
\begin{align}
  \lambda_m^*(t)\,dt 
  &=
  \lambda_m(t|\mcH_t) \, dt \\
  &=
  \Pr\left( t_{i}^m\in [t,t+dt)|\mcH_t\right),
\end{align}
which specifies the probability that a label $m$
occurs in the infinitesimal time interval
$[t, t+dt)$
given past events
$\mcH_t=\left\{t_i^m\in\mcH | t_i < t\right\}$.
We follow \citet{daleyIntroductionTheoryPoint2003} in using
$\lambda_m^*(t)=\lambda_m(t|\mcH_t)$
to indicate $\lambda_m^*(t)$ is conditioned on past events,
and $t_i^m$ to refer to a label of type $m$ at time $t_i$.
Given a specified conditional intensity $\lambda_m^*(t)$, the conditional density $p^*_m(t)$ is, for $t_{i-1}<t\leq t_{i}$,
\begin{align}
    \label{eq:conditional-density}
    p_m^*(t)
    &=
   \lambda_m^*(t)
    \,\exp\left[
    -\sum_{n=1}^M
    \Lambda_n^*(t)
    \right],
\end{align}
where $\Lambda_m^*(t)$ is the cumulative intensity:
\begin{align}
  \label{eq:cumulative-density}
    \Lambda_m^*(t)
    =
    \Lambda_m(t|\mcH_t)
    &=
    \int_{t_{i-1}}^t \lambda_m^*(t^\prime)\,dt^\prime.
\end{align}

In a multi-class setting, where \emph{exactly} one label (displayed as the indicator function $\mathbbm{1}$) is present in any event, the log-likelihood of a sequence $\mcH$ is a form of categorical cross-entropy:
\begin{multline}
\label{eq:log-likelihood-multi-class}
    \log p_{\textrm{multi-class}}(\mcH)
    =
    \underbrace{\sum_{m=1}^M
    \sum_{i=1}^N
    \mathbbm{1}_{i,m}\log p_m^*(t_{i})}_\text{events at $t_1, \dots, t_N$}\\
    -
    \underbrace{\sum_{m=1}^{M} \int_{t_N}^{w_+}\lambda_m^*(t^\prime)
    \,
    dt^\prime}_\text{no events between $t_N$ and $w_+$},
\end{multline}
where $\mcH_{t_0}=\mcH_{t_1}=\{\}$, and $t_0= w_-$ but does not correspond to an event.

In a multi-label setting, where \emph{at least} one label is present in any event, the log-likelihood of a sequence $\mcH$ is
a form of binary cross-entropy:
\begin{multline}
    \label{eq:log-likelihood-multi-label}
    \log p_{\textrm{multi-label}}(\mcH)
    =
    \log p_{\textrm{multi-class}}(\mcH) \\
    +
    \sum_{m=1}^M
    \sum_{i=1}^N
    (1- \mathbbm{1}_{i,m})
    \log \Big[1-p_m^*(t_{i})\Big].
\end{multline}
This setting should be especially useful to model \glspl{ehr}, as a single medical consultation usually includes various events, such as diagnoses or prescriptions, all happening at the same time.

\subsection{Neural temporal point processes}
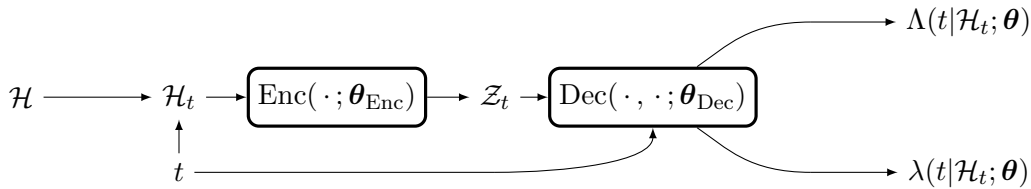
\begin{figure*}[!htb]
  \centering
   \begin{tikzpicture}[align=center,xscale=2.1]
     \node at        ( 0, 0) (t)  {$t$};
     \node at        (-1, 1) (h)  {$\mcH$};
     \node at        ( 0, 1) (ht) {$\mcH_t$};
     \node[punkt] at ( 1, 1) (enc) {$\text{Enc}(\,\cdot\,;\mbtheta_{\text{Enc}})$};
     \node        at ( 2, 1) (zt) {$\mcZ_t$};
     \node[punkt] at ( 3, 1) (dec) {$\text{Dec}(\,\cdot\,,\,\cdot\,;\mbtheta_{\text{Dec}})$};
     \node        at ( 5, 2) (Lam) {$\Lambda(t|\mcH_t;\mbtheta)$};
     \node        at ( 5, 0) (lam) {$\lambda(t|\mcH_t;\mbtheta)$};
    \path[->]
   (t)        edge                     node                {}      (ht)
   (h)        edge                     node                {}      (ht)
   (ht)       edge                     node                {}      (enc)
   (enc)      edge                     node                {}      (zt)
   (zt)       edge                     node                {}      (dec)
   (t.east)   edge[-]                  node                {}      (2,0)
   (2,0)      edge[out=0, in=270]      node                {}      (dec.south)
   (dec)      edge[-, out=45, in=180]  node                {}      (4,2)
   (4,2)      edge                     node                {}      (Lam)
   (dec)      edge[-, out=-45,in=180]  node                {}      (4,0)
   (4,0)      edge                     node                {}      (lam)
;
  \end{tikzpicture}
 \caption{
Encoder/decoder architecture of Neural \glspl{tpp}.
Given a query time $t$,
the sequence $\mcH$ is filtered to the events $\mcH_t$ in the past of $t$.
The encoder maps $\mcH_t$ to continuous
representations $\mcZ_t=\{\bmz_i\}_{i=1}^{|\mcH_t|}=\textrm{Enc}(\mcH_t;\mbtheta_{\textrm{Enc}})$.
Each $\bmz_i$ can be considered as a contextualised representation for the event at $t_i$.
Given $\mcZ_t$ and $t$, the decoder outputs $\textrm{Dec}(t,\mcZ_t;\mbtheta_{\textrm{Dec}}) \in \mbbR^M$
that the conditional intensity and conditional cumulative intensity are derived from without any learnable parameters.}
\label{fig:enc-dec}
\end{figure*}

Encoder-decoder architectures have proven effective for \gls{nlp}
\citep{choLearningPhraseRepresentations2014,kirosSkipThoughtVectors2015,hillLearningDistributedRepresentations2016,vaswaniAttentionAllYou2017}.
Existing Neural \glspl{tpp} also exhibit this structure: 
the encoder creates event representations based only on information about other events;
the decoder takes these representations and the decoding time to produce a new representation.
The output of the decoder produces the conditional intensity and conditional
cumulative intensity at that decoding time.
For more detail, see \Cref{fig:enc-dec}.

\section{Previous work}
While a Neural TPP encoder can be readily chosen from existing sequence models such as \glspl{rnn}, choosing a decoder is much less straightforward due to the integral in \Cref{eq:cumulative-density}.
Existing work can be categorised based on the relationship between the conditional intensity $\lambda^*(t)$ and conditional cumulative intensity $\Lambda^*(t)$. 
We focus on three approaches: 
\begin{itemize}
    \item Closed form likelihood: the conditional density $p^*(t)$ is closed form,
    \item Analytic conditional intensity: only $\lambda^*(t)$ is closed form, $\Lambda^*(t)$ is estimated numerically,
    \item Analytic cumulative conditional intensity: only $\Lambda^*(t)$ is closed form, $\lambda^*(t)$ is computed through differentiation.
\end{itemize}

\subsection{Closed form likelihood}
The closed form likelihood approach implies that the contribution from each event to the likelihood
$p^*_m(t)=\lambda_m^*(t)\exp[-\sum_{n=1}^M\Lambda_n^*(t)]$
is closed form.

A well-known example is the Hawkes process 
\citep{hawkesSpectraSelfexcitingMutually1971,nickelLearningMultivariateHawkes2020a}, 
which models the conditional intensity as
\begin{equation*}
 \lambda_m^*(t)
=\mu_m+\sum_{n=1}^M\alpha_{m,n}\sum_{i:t_i^n<t}\exp[-\beta_{m,n}(t-t_i^n)]   
\end{equation*}
with learnable parameters
$\mbmu\in\mbbR^{M}_{>0}$,
$\mbalpha\in\mbbR^{M\times M}_{\geq0}$, and
$\mbbeta\in\mbbR^{M\times M}$.
The closed form of
$\Lambda^*_m(t;\mbtheta)$ comes from the simple exponential linear $t$-dependence of
$\lambda_m^*(t;\mbtheta)$, which limits the model flexibility.

\citet{duRecurrentMarkedTemporal2016} leveraged the same closed form,
conditioning an exponential linear decoder on the output of a \gls{rnn} encoder (RMTPP).
As with the Hawkes process, the model can only model exponential dependence in time.
Additionally, it assumes labels are conditionally independent of time given a history, which we will show is a limiting assumption in domains like modelling \glspl{ehr}.


An alternative to taking 
$\lambda^*(t)$ and 
$\Lambda^*(t)$ closed form is to take
$p^*(t)$ closed form.
\citet{shchurIntensityFreeLearningTemporal2020} directly approximate the conditional density using a log-normal mixture (LNM).
With a sufficiently large mixture, $p^*(t)$ can approximate any conditional density and is a more flexible model than the RMTPP.
However, as in \citet{duRecurrentMarkedTemporal2016},
labels are modelled conditionally independent of decoding time.

While no decode-time dependent Neural \glspl{tpp} methods have been applied to modelling \glspl{ehr}, \citet{weiss2013forest} and \citet{lian2015multitask} apply conditional Poisson processes to various \glspl{ehr} datasets, \citet{islamMarkedPointProcess2017} combines a conditional Poisson process with a Gaussian distribution to predict in hospital mortality using ICU data, and \citet{zhang2020adverse} uses an individual heterogeneous conditional Poisson model on longitudinal \glspl{ehr} for adverse drug reaction discovery.

\subsection{Analytic conditional intensity}
The analytic conditional intensity approach 
approximates the conditional intensity with a \gls{nn} whose output
is positive
\begin{equation*}
 \lambda^*_m(t;\mbtheta)=\textrm{Dec}(t,\mcZ_t;\mbtheta_{\textrm{Dec}})_m\in\mbbR_{\geq0},   
\end{equation*}
and whose $t$-integral must be approximated numerically.

The positivity constraint is satisfied by requiring the final activation function to be positive. 
Early Neural \glspl{tpp} employed exponential activation \citep{duRecurrentMarkedTemporal2016}. 
Recently, the scaled softplus activation 
\begin{equation*}
   \sigma_+(x_m)=s_m \log(1+\exp(x_m/s_m))\in\mbbR_{>0} 
\end{equation*}
with learnable $\bms\in\mbbR^M$ has gained popularity \citep{meiNeuralHawkesProcess2017}. 
Ultimately, writing a neural approximator for $\lambda_m^*(t)$ is relatively simple as there is no constraint on the \gls{nn} architecture itself. Of course, this does not mean it is easy to train.

To approximate the integral, \gls{mc} estimation
can be employed
given a sampling strategy \citep{meiNeuralHawkesProcess2017}. \citet{zhuDeepAttentionSpatioTemporal2020} also applies this strategy for spatio-temporal point process, which jointly models times and labels, but only uses an embedding layer as an encoder.


\subsection{Analytic conditional cumulative intensity}
\label{sec:anayltic-cumulative}
The analytic conditional cumulative intensity approach approximates the 
conditional cumulative intensity with a \gls{nn}
whose output is positive
\begin{equation*}
   \Lambda^*_m(t;\mbtheta)=\textrm{Dec}(t,\mcZ_t;\mbtheta_{\textrm{Dec}})_m\in\mbbR_{\geq0} 
\end{equation*}
and whose derivative is positive and approximates the conditional intensity
\begin{equation*}
\lambda^*_m(t;\mbtheta)=d\textrm{Dec}(t,\mcZ_t;\mbtheta_{\textrm{Dec}})_m/dt\in\mbbR_{\geq0}
\end{equation*}
This derivative can be computed using backpropagation.
While the cumulative intensity approach removes the variance induced by \gls{mc} estimation, the monotonicity entails specific constraints on the \gls{nn}.

\citet{omiFullyNeuralNetwork2019} uses a \gls{mlp} with positive weights in order to model a monotonic decoder. It also uses tanh activation functions, and a softplus final activation function to ensure a positive output. However, this model does not handle labels, and has been criticised by \citet{shchurIntensityFreeLearningTemporal2020} for violating $ \Lim{t \to +\infty} \Lambda^*_m(t;\mbtheta) = +\infty $.


\section{Proposed models}
Our goal is to model the joint distribution of \glspl{ehr} which
contain long-term, non-sequential dependencies.
This makes the successful application of \glspl{rnn} challenging due to their recency bias
\citep{bahdanauNeuralMachineTranslation2016,ravfogelStudyingInductiveBiases2019}.
Additionally, \glspl{rnn} do not meet the explainability standard required for real-world application;
although there are methods for interpreting \gls{rnn} outputs \citep{samek2019towards}, 
when deployed in healthcare applications, models should be interpretable without relying on auxiliary models \citep{rudinStopExplainingBlack2019}.

To address these challenges, we propose the use of attention. 
This is capable of dealing with long range context \citep{liu2018generating}, and 
benefits some model explainability\footnote{This
intepretability may be limited to the importance of events in a patient record \citep{serrano-smith-2019-attention}, which without auxiliary models provides much clinical utility. For example, rapidly provisioning important information about a patient to a doctor to facilitate greater focus on patient need \citep{Overhage2020}.} 
without requiring auxiliary models.
In this section we develop the necessary building blocks to define attention-based Neural \glspl{tpp} that directly model either the conditional intensity
$\lambda_m^*(t)$
or the conditional cumulative intensity
$\Lambda_m^*(t)$, alongside several benchmark Neural \glspl{tpp}.
In total, we present 2 encoders (\Cref{subsec:encoders}) and 7 decoders (\Cref{subsec:decoders}) whose combination yields the 14 NeuralTPPs we evaluate\footnote{Our implementation of these models, as well as the experimental setup can be found at \url{https://github.com/babylonhealth/neuralTPPs}.}.

\subsection{Encoders}
\label{subsec:encoders}
As in RMTPPs \citep{duRecurrentMarkedTemporal2016} we use a \gls{gru} encoder (\textbf{GRU}).
We also evaluate a continuous-time self-attention encoder \citep{vaswaniAttentionAllYou2017,xiongLayerNormalizationTransformer2020} (\textbf{SA}).
Precise forms are presented in \Cref{app:taxonomy-encoders}.

\subsection{Decoders}
\label{subsec:decoders}
Precise forms for all decoders are presented in \Cref{app:taxonomy-decoders}.
\subsubsection{Closed form likelihood}
We take two decoders from the literature:
a conditional Poisson process (\textbf{CP}), a \textbf{RMTPP} \citep{duRecurrentMarkedTemporal2016}, and a log-normal mixture (\textbf{LNM})  \citep{shchurIntensityFreeLearningTemporal2020}.

\subsubsection{Analytic conditional intensity}
\label{subsec:analytic-cond-int}
To model a positive $\lambda_m^*(t)$ we follow \citet{meiNeuralHawkesProcess2017}: 
we apply a softplus activation to the final layer, and use \gls{mc} estimation with a single, uniformly distributed sample per time interval.

Using this strategy, we propose a decoder based on a MLP (\textbf{MLP-MC}), and an attention mechanism (\textbf{Attn-MC}).

\subsubsection{Analytic conditional cumulative intensity}
\label{subsec:positive-monotonic-approximators}
\begin{table*}[htbp]
  \caption{Properties of datasets used for evaluation.}
  \centering
  \scriptsize
  \begin{tabular}{lllllllll}
    \toprule
    &&&&& \multicolumn{4}{c}{Size} \\
    \cmidrule(r){6-9}
    Dataset       & \# classes   & Task type & \# events & Avg. length & Train & Valid & Test & Batch \\ 
    \midrule
    Hawkes (ind.)   & 2   & Multi-class        & 457,788   & 19          & 16,384     & 4,096    & 4,096    & 512 \\ 
    Hawkes (dep.)     & 2   & Multi-class        & 607,512   & 25          & 16,384     & 4,096    & 4,096    & 512 \\ 
    MIMIC-II       & 75  & Multi-class        & 2,419     & 4           & 585        & NA       & 65    & 65    \\ 
    Stack Overflow & 22  & Multi-class        & 480,413   & 72          & 5,307      & NA       & 1,326 & 32    \\
    Retweets       & 3   & Multi-label         & 2,087,866 & 104         & 16,000     & 2,000    & 2,000    & 256 \\ 
    Ear infection  & 39  & Multi-label         & 14,810    & 2           & 8,179      & 1,022    & 1,023    & 512 \\ 
    Synthea        & 357 & Multi-label         & 496,625   & 43          & 10,524     & 585      & 585 & 64      \\ 
    \bottomrule
  \end{tabular}
  \label{tab:data_stats}
\end{table*}

Finally, we consider modelling $\Lambda_m^*(t)=\textrm{Decoder}(t,\mcZ_t;\mbtheta)_m$. 
This is more difficult as the decoder must satisfy four properties:
\begin{multicols}{2}
\begin{enumerate}
    \item $\Lambda_m^*(t)>0$,
    \item $\Lambda_m^*(t_i)=0$,
    \item $\Lim{t\to\infty}\Lambda_m^*(t)=\infty$,
    \item $d\Lambda_m^*(t)/dt>0$.
\end{enumerate}
\end{multicols}
1. is solved using a final layer positive activation.
2. is solved by parameterising the decoder as $\textrm{Decoder}(t,\mcZ_t;\mbtheta)=f(t,\mcZ_t;\mbtheta)-f(t_i,\mcZ_t;\mbtheta)$.
3. can be satisfied in two ways.
The approach we take is to add a Poisson term to the conditional intensity (see \Cref{subsec:base-intensity}).
An alternative is to use a non-saturating activation function, which we investigate in \Cref{app:monotonic}.

4. is the most challenging.
If $f$ represents a $L$-layer \gls{nn}, where the output of each layer $f_{i}$ is fed into the next $f_{i+1}$, we can write $f(t)=(f_L \circ \cdots \circ f_1)(t)$. Providing
each step produces an output that is a monotonic in its input, i.e. $df_i/df_{i-1}\geq0$ and $df_1/dt\geq0$, then $f(t)$ is a monotonic function of $t$. 

Given that softmax, layer normalisation, trigonometric encoding, and projection are inconsistent with monotonicity, many modifications were required in order to use MLPs and attention networks for cumulative modelling. 
Given the the importance of $t$-derivatives, we use the adaptive Gumbel activation \citep{farhadiActivationAdaptationNeural2019}, allowing learnable adaption of first derivatives. 
All proposed modifications are presented in \Cref{app:monotonic}.

Using these modifications, we propose two cumulative decoders: a MLP (\textbf{MLP-CM}) and an attention mechanism (\textbf{Attn-CM}), which handle labels and address the criticisms of \citet{shchurIntensityFreeLearningTemporal2020}.


\subsection{Base intensity}
\label{subsec:base-intensity}
To every model\footnote{We do not add a Poisson term to the conditional Poisson process. Moreover, the original RMTPP does not include this Poisson term, however we include it to provide  a fairer comparison.} we add a Poisson process\footnote{In the \gls{tpp} literature, the Poisson term is often referred to as the exogenous intensity, and the time-dependent piece the endogenous impact.} 
\begin{equation}
    \bm\lambda^*_{\textrm{Total}}(t)=\alpha_1\,\bm\mu + \alpha_2\,\bm\lambda^*(t)
\end{equation}
where $\bm\alpha=\softmax(\bma)$, with $\bma$ learnable.

We found that initialising $\bma$ such that $\alpha_1\sim e^3\alpha_2$, i.e. starting the combined \gls{tpp} as mostly Poisson, aided convergence.

\section{Evaluation}

\subsection{Datasets}
Dataset statistics are summarised in \Cref{tab:data_stats}.
\paragraph{Hawkes Processes} This synthetic data allows us to have access to a
theoretically infinite amount of data. Moreover, as we know the intensity
function, we were able to compare each of our modelled
intensity against the true one. 
We designed one dataset consisting of two independent processes, \textbf{Hawkes (independent)}, and a second one allowing interactions between two processes, \textbf{Hawkes (dependent)}.

\paragraph{Baselines} We also compared our models on datasets commonly used to evaluate \glspl{tpp}. We use:
\textbf{MIMIC-II}\footnote{Although this dataset contains EHRs, they are not longitudinal. We only include it as a known \gls{tpp} benchmark.}, a medical dataset of clinical visits to Intensive Care Units,
\textbf{Stack Overflow}, which classifies users on this question answering
website, and
\textbf{Retweets}, which consists in streams of retweet events by different
types of Twitter users.
Further details about the datasets and their preprocessing can be found in
\citet{duRecurrentMarkedTemporal2016,meiNeuralHawkesProcess2017}.

\paragraph{Synthetic \glspl{ehr}} We used the Synthea simulator \citep{walonoskiSyntheaApproachMethod2018} which generates patient-level \glspl{ehr} using human expert curated Markov processes.

We created a dataset utilising all of Synthea's modules, \textbf{Synthea (Full)}, as well as a dataset generated solely from the ear infection module \textbf{Synthea (Ear Infection)}\footnote{Ear infection was selected for a simpler benchmark as the generation process demonstrated temporal dependence}. For each task, we generated approximately 10,000 patients, and we kept 10\% of the data for validation and testing, using 5 different folds randomly chosen.
Moreover, to optimise GPU utilisation, we truncated sequences larger than 400 events, which amounts to around 0.4\% of the dataset.

We filtered out all events except conditions and medications.
For each patient, we transformed event times by subtracting their birth date, then multiplying by $10^{-5}$.
When provided, their death date is used as the end of the observation window $w_+$, otherwise we use the latest event in their history.

\subsection{Hyperparameters}
Encoder and decoder were set to one layer each. 
The size of each layer was 8 for the small datasets (Hawkes and MIMIC-II), 32 for the medium ones (Ear infection, Stack Overflow and Retweets), and 64 for Synthea. 
Batch size adjusted per-dataset due to GPU memory availability (see \Cref{tab:data_stats}).
The Adam optimizer \citep{kingmaAdamMethodStochastic2017} was used with the Noam learning rate scheduler \citep{vaswaniAttentionAllYou2017}. 
The scheduler had a maximum learning rate value of 0.01 and 10 epochs of warming-up, allowing the Poisson component to be adjusted before tuning the temporal piece. 
All our models were trained using \gls{mle}, and early stopping was employed with a patience of 100 epochs.

\subsection{Metrics}
\begin{table*}[!htb]
  \caption{Evaluation on multi-class tasks. On both tables, ($\dagger$) indicates a model newly presented in this work. Best performance and performances whose confidence intervals overlap the best are boldened.}
  \centering
  \tiny
  \scriptsize
  \begin{tabular}{lllllll}
    \toprule
    Encoder GRU & \multicolumn{2}{c}{MIMIC-II} & \multicolumn{2}{c}{Stack Overflow} & Hawkes (ind.) & Hawkes (dep.) \\
    \cmidrule(r){2-3}\cmidrule(r){4-5}
    Decoder & F1 score & NLL/time    & F1 score & NLL/time & NLL/time & NLL/time \\ 
    \midrule
    CP & \textbf{.691 (.083)} & 6.78 (1.99) & .325 (.004) & .553 (.003) & .623 (.002) & .774 (.004)  \\
    RMTPP        & .215 (.12)  & 11.4 (3.57) & .284 (.005) & .592 (.006) & .610 (.004) & .731 (.004) \\ 
    LNM          & \textbf{.705 (.17)}  & 6.33 (.37)  & .314 (.003) & .548 (.004) & \textbf{.605 (.001)} & \textbf{.727 (.001)} \\ 
    MLP-CM$^\dagger$       & .166 (.083) & 12.2 (3.57) & .305 (.016) & .573 (.004) & .609 (.005) & \textbf{.728 (.003)} \\ 
    MLP-MC$^\dagger$       & \textbf{.665 (.050)} & 9.45 (3.87) & \textbf{.333 (.008)} & \textbf{.540 (.006)} & \textbf{.605 (.001)} & \textbf{.727 (.001)} \\
    Attn-CM$^\dagger$        & .189 (.10)  & 10.4 (2.34) & .286 (.017) & .589 (.012) & .613 (.003) & .732 (.002) \\ 
    Attn-MC$^\dagger$        & \textbf{.641 (.131)}  & \textbf{5.27 (2.05)} & .314 (.006) & .560 (.005) & .612 (.005) & .729 (.002) \\ 
    \hline
    Encoder SA &&&&&& \\
    \hline
    CP & \textbf{.686 (.067)} & 7.26 (1.44) & .326 (.003) & .555 (.003) & .622 (.001) & .774 (.002) \\ 
    RMTPP        & \textbf{.709 (.076)} & \textbf{4.24 (2.66)} & .288 (.002) & .592 (.002) & .609 (.005) & .734 (.002) \\ 
    LNM          & \textbf{.632 (.20)}  & 7.76 (2.42) & .305 (.003) & .561 (.006) & \textbf{.607 (.005)} & .730 (.004) \\ 
    MLP-CM$^\dagger$       & .159 (.088) & 12.7 (3.45) & .292 (.003) & .634 (.042) & .611 (.005) & .733 (.002) \\ 
    MLP-MC$^\dagger$       & .498 (.150) & 10.82 (2.13) & .327 (.016) & .557 (.015) & \textbf{.606 (.002)} & .730 (.001) \\
    Attn-CM$^\dagger$        & .186 (.082) & 11.4 (3.90) & .269 (.011) & .665 (.047) & .614 (.001) & .733 (.001) \\ 
    Attn-MC$^\dagger$        & \textbf{.648 (.098)} & \textbf{4.61 (2.49)} & \textbf{.342 (.006)} & \textbf{.543 (.005)} & \textbf{.605 (.001)} & \textbf{.728 (.001)} \\ 
    \bottomrule
  \end{tabular}
  \label{tab:multi-class}
\end{table*}

\begin{figure*}[!htb]
\small
\floatconts
  {fig:synthea-intensity}
  {\caption{Intensity functions on several labels applied to the same \gls{ehr}. We can see that the intensity of the GRU-Attn-MC decreases more smoothly than the GRU-CP on the Otitis media label and is able to spike on the recurring medicine prescription. This enables the GRU-Attn-MC to achieve a lower NLL than the GRU-CP.}}
  {%
    \subfigure[GRU-Conditional-Poisson]{\label{fig:intensity-first}%
      \includegraphics[width=0.5\linewidth]{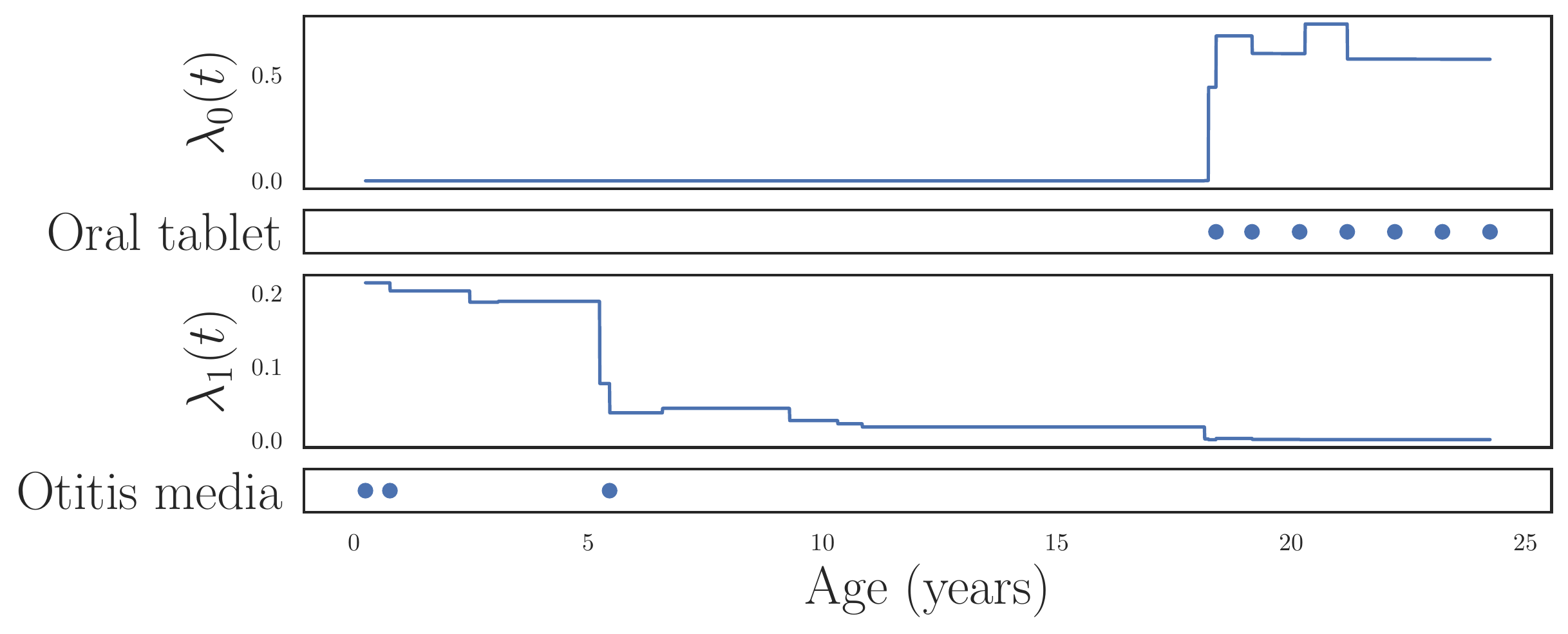}}%
    \subfigure[GRU-Attn-MC]{\label{fig:intensity-second}%
      \includegraphics[width=0.5\linewidth]{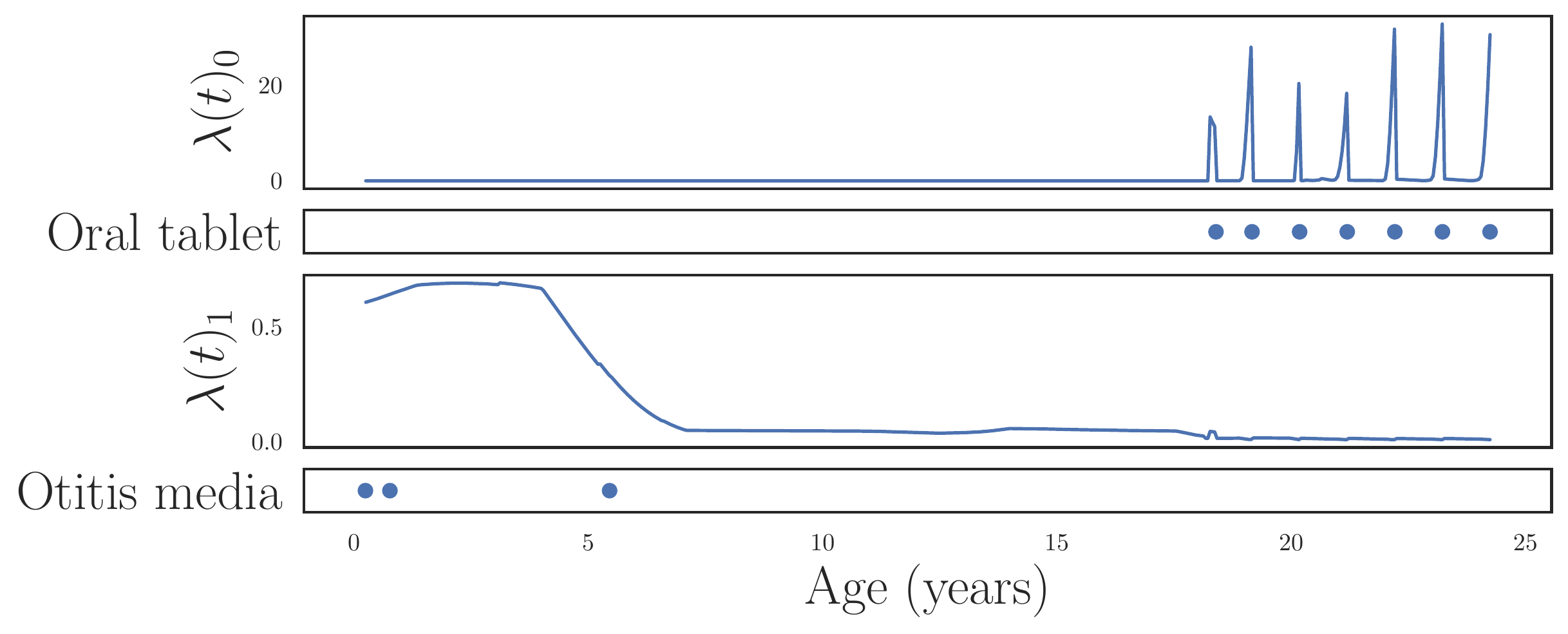}}
  }
\end{figure*}

Models were evaluated using a weighted F1 metric for multi-class datasets, and a weighted ROC-AUC metric for the multi-label ones. This enabled us to assess a model's ability predict the label(s) of the next event given the time it occurred and the previous events.

\gls{nll} normalised by time\footnote{Time normalisation makes the metric meaningful over sequences defined on different time intervals.} (\gls{nll}/time) was also used to evaluate models. 
This metric enables us to assess a model's ability to predict the label(s) of the next event, as well as when that event will happen given the previous events.

We want to stress that neither of these metrics is a perfect measure of performance in isolation.
Indeed, F1 and ROC-AUC reward a model for accurate predictions given an event instance, but do not penalise a model for inaccurate predictions in the inter-event interval. 
This happens in \Cref{fig:synthea-intensity}.

\begin{table*}[!htb]
  \caption{Evaluation on multi-label tasks.}
  \centering
  \tiny
  \scriptsize
  \begin{tabular}{lllllll}
    \toprule
    Encoder GRU & \multicolumn{2}{c}{Retweets} & \multicolumn{2}{c}{Synthea (Ear infection)} & \multicolumn{2}{c}{Synthea (Full)} \\
    \cmidrule(r){2-3}\cmidrule(r){4-5}\cmidrule(r){6-7}
    Decoder & ROC-AUC & NLL/time    & ROC-AUC & NLL/time & ROC-AUC & NLL/time \\ 
    \midrule
    CP & \textbf{.611 (.001)} & 5.71 (1.02)  & .786 (.015) & 37.6 (7.76) & \textbf{.850 (.014)} & 89.6 (3.15) \\
    RMTPP        & .532 (.003) & -9.05 (1.40) & .672 (.013) & 36.7 (4.25) & .616 (.043) & 62.2 (6.63) \\ 
    LNM          & .521 (.010) & \textbf{-10.9 (2.11)} & .765 (.008) & \textbf{26.7 (5.35)} & .770 (.010) & \textbf{9.49 (6.37)} \\ 
    MLP-CM$^\dagger$       & .533 (.001) & \textbf{-10.4 (.301)} & .742 (.050) & 30.4 (7.03) & .692 (.051) & 34.2 (14.5) \\ 
    MLP-MC$^\dagger$       & .535 (.001) & \textbf{-10.3 (.186)}  & .838 (.024) & \textbf{29.2 (9.09)} & .789 (.024) & \textbf{15.7 (9.57)} \\
    Attn-CM$^\dagger$        & .526 (.001) & \textbf{-10.0 (.535)} & .799 (.028) & \textbf{27.1 (4.95)} & .508 (.000) & 80.9 (2.51) \\ 
    Attn-MC$^\dagger$        & .532 (.001) & -9.14 (.358) & \textbf{.857 (.005)} & \textbf{25.1 (4.45)} & .822 (.006) & 21.9 (2.05) \\ 
    \hline
    Encoder SA &&&&&& \\
    \hline
    CP & .608 (.001) & 4.85 (.021)  & .792 (.009) & 38.6 (8.96) & .825 (.020) & 91.2 (4.24) \\ 
    RMTPP        & .535 (.001) & -8.70 (.087) & .675 (.068) & 42.9 (9.01) & .593 (.029) & 67.8 (6.13) \\ 
    LNM          & .528 (.007) & \textbf{-10.9 (1.29)} & .767 (.007) & \textbf{26.2 (5.51)} & .760 (.014) & 27.5 (27.2) \\ 
    MLP-CM$^\dagger$       & .512 (.001) & -8.93 (.180) & .684 (.094) & 33.8 (15.3) & .589 (.123) & 62.8 (30.8) \\ 
    MLP-MC$^\dagger$       & .533 (.001) & -9.21 (.174) & \textbf{.849 (.016)} & \textbf{24.9 (5.10)} & .796 (.007) & 18.5 (1.20) \\
    Attn-CM$^\dagger$        & .517 (.001) & -8.54 (.168) & .697 (.059) & 31.4 (6.90) & .503 (.004) & 85.0 (3.01) \\ 
    Attn-MC$^\dagger$        & .534 (.001) & -9.48 (.188) & \textbf{.855 (.010)} & \textbf{24.5 (4.56)} & .805 (.007) & 27.1 (4.77) \\ 
    \bottomrule
  \end{tabular}
  \label{tab:multi-label}
\end{table*}

Alternatively, the \gls{nll} is prone to overfit: if a Neural \gls{tpp} accurately guesses when an event will happen, it has the incentive to endlessly increase the intensity at that time, reducing the overall NLL without improving the resulting model meaningfully. Ideally, a model should perform well on both metrics.

\section{Results}

We report our results on multi-class tasks (\Cref{tab:multi-class}) and on multi-labels tasks (\Cref{tab:multi-label}).

First, we note that the GRU-CP has a competitive NLL and F1 score on the MIMIC-II and StackOverflow datasets.
As the the CP decoder is decode time-independent, the result indicates that MIMIC-II and StackOverflow may not be appropriate for benchmarking \glspl{tpp}. We suggest caution for future work using these datasets.
All other datasets appear suitable. We focus here on \glspl{ehr}, and discuss the Hawkes and Retweets results in \Cref{app:additional-results}.

On \glspl{ehr}, we see that modelling time and labels jointly yields better results than assuming them independent given the history. 
Indeed, most of our models outperform the RMTPP on all metrics, and, while the LNM is the best in terms of NLL, it achieves a significantly lower ROC-AUC score than all of our models except those using an Attn-CM decoder.
Overall, our MC-based models outperform the CP in terms of NLL/time, and the LNM in terms of ROC-AUC.

Despite Attn-MC decoders performing strongly across all baseline datasets, as well as the Synthea (Ear Infection), this performance gap is not as wide on Synthea (Full). 
Studies have found Synthea to be medically representative \citep{walonoskiSyntheaApproachMethod2018,Chen2019}. 
However,
it is possible that noise in Synthea (Full) may be obfuscating the temporal progression of diseases (non-clinical investigation of the records showed 36.5\% received a diagnosis of prediabetes after being diagnosed with diabetes).
Applying Neural \gls{tpp} to genuine EHRs would be a fruitful line of future enquiry.

Although the cumulative-based models have theoretical benefits compared with their Monte Carlo counterparts, they are in practice harder to train. This may explain their relatively poor performance on datasets such as on MIMIC-II and Synthea (Full), and the high variance between F1/ROC-AUC scores on these datasets.

\section{Conclusion}
In this work we gathered and proposed several neural network parameterisations of \glspl{tpp}, evaluating them on synthetic \glspl{ehr}, as well as common benchmarks.

Given the significant out-performance of our models on synthetic \glspl{ehr}, labels should be modelled jointly with time, rather than be treated as conditionally independent; a common simplification in the TPP literature. 
We also note that common TPP metrics are not good performance indicators on their own. For a fair comparison we recommend using multiple metrics, each capturing distinct performance characteristics.

By employing a simple test that checks a dataset's validity for benchmarking TPPs, we demonstrated potential issues within several widely-used benchmark datasets.
We recommend caution when using those datasets, and that our test be run as a sanity check of any new evaluation task.

Finally, we have demonstrated that attention-based TPPs appear to transmit pertinent EHR information and perform favourably compared to existing models. This is an exciting line for further enquiry in EHR modelling, where human interpretability is essential. Future work should apply these models to real EHR data to investigate medical relevance.

\section*{Acknowledgements}
We are grateful to Sheldon Hall, Jeremie Vallee and Max Wasylow for assistance with experimental infrastructure, and to Kristian Boda, Sunir Gohil, Kostis Gourgoulias, Claudia Schulz and Vitalii Zhelezniak for fruitful comments and discussion.

\bibliography{main}

\appendix
\appendix
\section{Positive monotonic building blocks}
\label{app:monotonic}

As discussed in 
\Cref{subsec:positive-monotonic-approximators},
if $f(t)$ is given by a $L$-layer \gls{nn}, where the output of each layer $f_{i}$ is fed into the next $f_{i+1}$, then we can write $f(t)=(f_L \circ \cdots \circ f_2 \circ f_1)(t)$. 
Application of the chain rule
\begin{equation}
\frac{df(t)}{dt}
  =
  \frac{df_L}{df_{L-1}}
  \frac{df_{L-1}}{df_{L-2}}
  \cdots
  \frac{df_2}{df_1}
  \frac{df_1}{dt}
\end{equation}
shows that a sufficient solution to achieving $df(t)/dt\geq0$ is to enforce
each step of processing to be a monotonic function of its input, i.e. $df_i/df_{i-1}\geq0$ and $df_1/dt\geq0$ implies $f(t)$ is a monotonic function of $t$.

As well as the monotonicity constraint, as we are interested in intensities that can decay as well as increase, it needs to be possible that the second derivative of $f(t)$ can be negative
\begin{equation}
  \frac{d\lambda(t|\mcH_t)}{dt} 
  \sim \frac{d^2f(t|\mcH_t)}{dt^2}
  <0
\end{equation}
for some history $\mcH_t$.

Here we collect all of the modifications we employ to ensure our architectures comply with this restriction.
\subsection{Linear projections}
\label{app:linear-monotonic}
Linear projections are a core component of dense layers, and can be made monotonic by requiring every element of the projection matrix is at least 0
\begin{align}
  f(\bmx)
  &=\bmW \bmx,\\
  df(\bmx)_i/dx_j=W_{i,j} 
  &\geq 0 \; \forall \; i,j.
\end{align}
One solution to this problem is to parameterise the $\bmW$ as a positive $g$ transformation of some auxiliary parameters $\bmV$ such that $\bmW=g(\bmV)\geq0$. 
Taking $g$ as \gls{relu}
leads to issue during training: any weight which reaches zero will become frozen no longer be updated by the network, harming convergence properties.
We also experimented using softplus and sigmoid activations.
These have the issue where the auxiliary weights $\bmV$ can be pushed arbitrarily negative and, when the network needs them again, training needs to pull them very far back for little change in $\bmW$, again harming convergence. 
This approach is also burdened with additional computational cost.

The most effective approach, which we employ in our models, involves modifying the forward pass such that any weights below some small $\epsilon>0$ are set to $\epsilon$.
This $\epsilon$ can be thought of as a lower bound on the derivative, has no computational overheads, and none of the issues discussed in the approaches above.
In our experiments we take $\epsilon=10^{-30}$.

\subsection{Dense activation functions}
\paragraph{ReLU}
The widely used \gls{relu} activation clearly satisfies the monotonicity constraint, however
\begin{align*}
    \frac{d\relu(x)}{dx}
    &=
    \begin{cases} 
    0 \quad \textrm{if} \quad x \leq 0 \\
    1 \quad \textrm{if} \quad x > 0,
    \end{cases}
    & \\
    \frac{d^2\relu(x)}{dx^2}
    &=0.
\end{align*}
As the second derivative is 0 then
\begin{equation}
 \frac{d\lambda_m^*(t)}{dt} = 0   
\end{equation}
which means that any cumulative model using the \gls{relu} activation is
equivalent to the conditional Poisson process
$\bm\lambda(\tau|\mcH_t)=\bm\mu(\mcH_t)$ introduced in \Cref{subsec:positive-monotonic-approximators}.
We cannot use \gls{relu} in a cumulative model.

\paragraph{Tanh}
\citet{omiFullyNeuralNetwork2019} proposed to use the $\tanh$ activation function
which doesn't have the constraints of \gls{relu}:
\begin{align*}
    \frac{d\tanh(x)}{dx}
    &={\sech}^2(x)\in(0,1),
    & \\
    \frac{d^2\tanh(x)}{dx^2}
    &=2\tanh(x)\,{\sech}^2(x)\in (c_-,c_+)
\end{align*}
where $c_{\pm}=\log(2\pm\sqrt2)/2$, so
$\tanh$ meets our requirements of a
positive first derivative and a non-zero second derivative.

\paragraph{Gumbel}
An alternative to $\tanh$ is the adaptive Gumbel activation introduced in
\citet{farhadiActivationAdaptationNeural2019}:
\begin{equation}
\label{gumbel}
  \sigma(x_m)
  =
  1-\left[ 1+s_m \exp(x_m) \right]^{-\frac1{s_m}}
\end{equation}
where
$\forall m:s_m\in\mbbR_{>0}$,
and $m$ is a dimension/activation index.
For brevity, we will refer to this activation function as the Gumbel activation, and
while its analytic properties we will
drop the dimension index $m$, however, since the activation is applied
element-wise, the analytic properties discussed directly transfer to the vector
application case.

Its first and second derivatives match our positivity and negative requirements
\begin{align}
  \label{eq:gumbel-first-derivative}
  \frac{d\sigma(x)}{dx}
    &=\exp(x) \nonumber\\ &\qquad \times \left[1+s\exp(x)\right]^{-\frac{s+1}s} \in (0,1/e),\\
    \label{eq:gumbel-second-derivative}
    \frac{d^2\sigma(x)}{dx^2}
    &=\exp(x)\left[1-\exp(x)\right] \nonumber\\ &\quad \times \left[1+s\exp(x)\right]^{-\frac{2s+1}s} \in (c_-,c_+),
\end{align}
where $c_{\pm} = (\pm\sqrt5 - 2)\exp[(\pm\sqrt5-3)/2]$.
The Gumbel activation shares many of the limiting properties of the $\tanh$ activation.
\begin{align}
\lim_{x\to-\infty}\sigma(x)&=0,
& \\
\lim_{x\to\infty}\sigma(x)
&=1,& \\
    \lim_{x\to\pm\infty}
    \frac{d\sigma(x)}{dx}
    &=
    \lim_{x\to\pm\infty}
    \frac{d^2\sigma(x)}{dx^2}
    =0.
\end{align}

For our purposes, the key advantage of the Gumbel over $\tanh$ are the learnable parameters $s_m$.
These parameters control the magnitude of the gradient through equation
\Cref{eq:gumbel-first-derivative}
and the magnitude of the second derivative through
\Cref{eq:gumbel-second-derivative}.
The maximum value of the first
derivative is obtained at $x=0$, which corresponds to the mode of the Gumbel
distribution, and 
for any value of $s$
\begin{equation}
  \max_x \frac{d\sigma(x)}{dx}=(1+s)^{-\frac{2s+1}{s}}.
\end{equation}
By sending $s\rightarrow0$ we get the largest maximum gradient of $1/e$ which
occurs in a short window in $x$, and by
sending $s\rightarrow\infty$ we get the smallest maximum gradient of $0$, which
occurs for all $x$.
This allows the activation function to control its sensitivity changes in the input, and allows the
\gls{nn} to be selective in where it wants fine-grained control over its
first and second derivatives
gradient (i.e. produce  therefore have an output that changes slowly over a large range in
the input values),
and where it needs the gradient to be very large in a small region of the input.

These properties are extremely beneficial for cumulative modelling, and we employ the Gumbel activation in all of these models.

\paragraph{Gumbel-Softplus}
Although $\tanh$ and adaptive Gumbel meet our positivity and negativity requirements, they share a further issue raised by 
\citet{shchurIntensityFreeLearningTemporal2020} in that their saturation ($\lim_{x\to\infty}\sigma(x)=1$)
does not allow the property $\lim_{t\to\infty} \Lambda_m^*(t) = \infty$, leading to an ill-defined joint probability density $p^*_m(t)$.
To solve this, we introduce the Gumbel-Softplus activation: \footnote{This modification could equally be applied to the $\tanh$ activation.}
\begin{equation}
  \sigma(x_m)
  =
  \text{gumbel}(x_m) \left[1 + \text{softplus}(x_m)\right],
\end{equation}
where gumbel is defined in \Cref{gumbel}, and softplus is the parametric softplus function:
\begin{equation}
  \text{softplus}(x_m)
  =
  \frac{\log\left(1 + s_m\,\exp(x_m)\right)}{s_m}
\end{equation}
This activation function has the property $\lim_{t\to\infty} \sigma(t) =
\infty$, in addition to satisfying the positivity and negativity constraints.

\subsection{Attention activation functions}
In addition to the activation functions used in the dense layers of \glspl{nn}, the
attention block requires another type of activation. 
This is indicated in
\Cref{attn}, as the attention logits $E_{i,j}$ are passed through a
function $g$ to produce the attention coefficients $\alpha_{i,j}$. 
Generally, this
activation function is the Softmax function (see \Cref{attn_activation}), however, this
function is not monotonic:
\begin{multline}
    \frac{\partial\softmax(x_i)}{\partial x_j} = \softmax(x_i) \\ * \left[\delta_{ij} - \softmax(x_j)\right] <0 \; \forall \; i\neq j,
    \label{eq:softmax-derivative}
\end{multline}
where $\delta_{ij}$ is the Kronecker delta function.

Consequently, we chose to use the sigmoid activation function instead of the
softmax when the modelled function required to be monotonic. Indeed, the sigmoid
function is monotonic:
\begin{equation}
    \frac{\partial\sigma(x)}{\partial x} = \sigma(x) \left[1 - \sigma(x)\right]>0.
\end{equation}
The softmax activation function has the nice property of making the attention
coefficient sum to one, therefore forcing the network to attend to a limited
number of points.
This can make the interpretation of the attention coefficient
easier.
However, using a softmax could potentially lead to a decrease of
performance if the points shouldn't strictly compete for contribution to the intensity under the generating process.
As a result, we believe that choosing a sigmoid or a softmax is not
straightforward.
We use sigmoid for cumulative approximators, and softmax everywhere else.

\subsection{Layer normalisation}
The final layer of the Transformer requiring modification issue is layer normalisation, a layer which dramatically improves the convergence speed of these models.

Modifying this layer requires realising that any layer requiring the division by the sum of activations (for example, even $L_2$ normalisation) will result in a negative derivative occurring somewhere.
Indeed, this is the underlying reason for the negative elements of the softmax Jacobian in \Cref{eq:softmax-derivative}.

It follows that any kind of normalisation cannot explicitly depend on the current set of activations.
Taking inspiration from batch normalisation, we construct an exponential moving average form of Layer normalisation during training.
When performing a forward pass, the means and standard deviations are taken from these moving averages and are treated as constants.
After the forward pass, the means and standard deviations are then updated in a similar fashion to batch normalisation.
Finally, taking the gain parameter positive as discussed in \Cref{app:linear-monotonic} results in a monotonic form of layer normalisation.

We employ this form of layer normalisation in the cumulative self-attention decoder (SA-CM), otherwise we use the standard form.

\subsection{Encoding}
\paragraph{Temporal Encoding for monotonic approximators}
The temporal embedding of \Cref{eq:temporal-embedding} is not monotonic in $t$ and
therefore cannot be used in any conditional cumulative approximator.
\citep{vaswaniAttentionAllYou2017} noted that a learnable temporal encoding had
similar performance to the one presented in \Cref{eq:temporal-encoding}.
In order to model the conditional cumulative intensity $\Lambda_m^*(t)$,
we will instead use a \gls{mlp}
\begin{multline}
  \label{eq:parametric-temporal-embedding}
  \textrm{ParametricTemporal}(t)=\\\textrm{MLP}(t;\mbtheta_{\geq\epsilon})\in\mbbR^{d_\textrm{Model}},
\end{multline}
where $\mbtheta_{\geq\epsilon}$ indicates that all projection matrices have positive values in all entries. Biases may be negative.
If we choose monotonic activation functions for the \gls{mlp}, then it
is a monotonic function approximator \citep{sillMonotonicNetworks1998}.

The $d_{\textrm{Model}}$ dimensional temporal encoding of an event at $t_i$ with labels $\mcM_i$ is then
\begin{align}
  \label{eq:temporal-encoding-app}
  \bmx_i
  &= \textrm{TemporalEncode}(t_i,\mcM_i) 
  \\
  &= 
  \bmv_i(\mcM_i)\sqrt{d_{\textrm{Model}}} +\nonumber\\
  &\qquad\textrm{ParametricTemporal}(t_i).
\end{align}

\subsection{Double backward trick}

When using a marked neural network based on the cumulative intensity, one issue comes from the fact that neural networks accumulate gradients. 
Specifically, the way that autograd is implemented in commonly used deep learning frameworks means that it is not possible to compute  $\frac{\partial}{\partial t} \Lambda_m^*(t)$ for a single $m$.
The value of the derivative will always be accompanied by the derivatives of all $m$ due to gradient accumulation, i.e. one can only compute the sum $\sum_{m=1}^M \frac{\partial}{\partial t} \Lambda_m^*(t)$ but not an individual $\frac{\partial}{\partial t} \Lambda_m^*(t)$.

A way of solving this issue is to split $\Lambda_m^*(t)$ into individual $M$ components, and to compute the gradient for each of them. However, this method is not applicable to high values of $M$ due to computational overhead.
Another way to solve this issue is to use the \emph{double backward trick} which allows to compute this gradient without having to split the cumulative intensity.

This trick is based on the following: we define a $B\times L\times M$ (Batch size $\times$ Sequence length $\times$ \# classes) tensor $\bma$ filled with zeros, and  compute the Jacobian vector product
\begin{equation}
    \textbf{jvp} = \left( \frac{\partial}{\partial t} \Lambda_m^*(t) \right)^T \bma \in\mbbR^{B\times L}.
\end{equation}
We then define second tensor $\bmb$ of shape $B\times L\times M$, filled with ones, and we compute the following Jacobian vector product:
\begin{align}
\left(\frac{\partial}{\partial \bma} \textbf{jvp} \right)^T &=
\left\{\frac{\partial}{\partial \bma} \left[ \left( \frac{\partial}{\partial t} \Lambda_m^*(t) \right)^T \textbf{a} \right] \right\}^T \\
\left(\frac{\partial}{\partial \bma} \textbf{jvp} \right)^T \bmb
&= \left( \frac{\partial}{\partial t} \Lambda_m^*(t) \right)^T \bmb \\
&= \frac{\partial}{\partial t} \Lambda_m^*(t)\in\mbbR^{B\times L\times M}
\end{align}
This recovers the element-wise derivative of the cumulative intensity function as desired, with the number of derivative operations performed being independent of the number of labels $M$.

\subsection{Modelling the log cumulative intensity}
We also investigated an alternative which is to let the decoder directly approximate  $ \log\Lambda_m^*(t)$.
We have then:
\begin{align}
    \frac{\partial \log \Lambda_m^*(t)}{\partial t} &= \frac{\lambda_m^*(t)}{\Lambda_m^*(t)},\\
    \log \lambda_m^*(t) &= \log \Lambda_m^*(t) \nonumber\\ &\quad
    + \log \left( \frac{\partial \log \Lambda_m^*(t)}{\partial t} \right).
\end{align}
While we didn't use this method to produce our results, we implemented it in our code-base. The advantage of this method is that the log intensity is directly modelled by the network, and its disadvantage is that the subtraction of the exponential terms to compute $\Lambda_m^*(t) - \Lambda_m^*(t_i)$ can lead to numerical instability.

\section{Taxonomy}
\label{app:taxonomy}

We ran our experiments using 2 encoders: SA and GRU, and 7 decoders: CP, RMTPP, LNM, MLP-CM, MLP-MC, Attn-CM, Attn-MM. We combined these encoders and decoders to form models, which are defined by linking an encoder with a decoder: for instance, GRU-MLP-CM denotes a GRU encoder with a MLP-CM (for cumulative) decoder. 
We present here these different components.

For the implementations of these models, as well as instances trained on the tasks in our setup, please refer to code base supplied in the supplementary material.

\paragraph{Label embeddings}
Although Neural \gls{tpp} encoders differ in how
they encode temporal information, they share a label embedding step.
Given labels $m\in\mcM_i$ localised at time $t_i$, the $\demb$
dimensional embedding $\bmv_i$ is
\begin{equation}
  \bmv_i = f_{\textrm{pool}} \left( \mcW_i=\{\bmw^{(m)} | m \in \mcM_i\} \right) \in \mbbR^{\demb}
\end{equation}
where $\bmw^{(m)}$ is the learnable embedding for class $m$, and
$f_{\textrm{pool}}(\mcW)$
is a pooling function, e.g.
mean pooling:
$f_{\textrm{pool}}(\mcW)=
\sum_{\bmw\in\mcW}\bmw/|\mcW|$, or
max pooling:
$f_{\textrm{pool}}(\mcW)=
\oplus_{\alpha=1}^{\demb}
\max(\{w_\alpha|\bmw\in\mcW\})$. \footnote{
In the multi-class setting, only one label appears at each time $t_i$, and so
$\bmv_i$ is directly the embedding for that label, and pooling has no effect.}

\subsection{Encoders}
\label{app:taxonomy-encoders}

\paragraph{GRU}
We follow the default equations of the GRU \gls{nn}:
\begin{align}
    \bmr_i &= \sigma(\bmW^{(1)} \bmx_i + \bmb^{(1)} + \\ \nonumber &\qquad \bmW^{(2)} \bmh_{(i-1)} + \bmb^{(2)}) \\
    \bmz_i &= \sigma(\bmW^{(3)} \bmx_t + \bmb^{(3)} +\nonumber \\ &\qquad \bmW^{(4)} \bmh_{(i-1)} + \bmb^{(4)}) \\
    \bmn_i &= \tanh(\bmW^{(5)} \bmx_i + \bmb^{(5)} + \nonumber\\ &\qquad \bmr_i \circ (\bmW^{(6)} \bmh_{(i-1)}+ \bmb^{(6)})) \\
    \bmh_i &= (1 - \bmz_i) \circ \bmn_i + \bmz_i \circ \bmh_{(i-1)},
\end{align}
where $\bmh_t$, $\bmr_t$, $\bmz_t$ and $\bmn_t$ are the hidden state, and the reset, update and new gates, respectively, at time t. $\bmx_t$ is the input at time $t$, defined by \Cref{eq:temporal-embedding}. $\sigma$ designs the sigmoid function and $\circ$ is the Hadamard product.

\paragraph{Self-attention (SA)}
We follow \Cref{attn} and \Cref{attn_activation} to form our attention block. The keys, queries and values are linear projections of the input: $\bmq = \bmW_q \bmx$, $\bmk = \bmW_k \bmx$, $\bmv = \bmW_v \bmx$, with the input $\mcX=\{\bmx_1,\bmx_2,\ldots,\}$, with each $x_i$ defined by \Cref{eq:temporal-encoding}.
We also apply a Layer Normalisation layer before each attention and feedforward block as in \citet{xiongLayerNormalizationTransformer2020}:
\begin{align}
    \mcQ^\prime &= \text{LayerNorm}(\mcQ), \\
    \mcQ^\prime &= \text{MultiHeadAttn}(\mcQ^\prime, \mcV) \\
    \mcQ^\prime &= \mcQ^\prime + \mcQ \\
    \mcZ &= \text{LayerNorm}(\mcQ^\prime) \\
    \mcZ &= \{\bmW^{(2)} \text{\gls{relu}}(\bmW^{(1)} \bmz_i + \bmb^{(1)}) + \bmb^{(2)}\} \\
    \mcZ &= \mcZ + \mcQ^\prime,
\end{align}
where MultiHeadAttn is defined by \Cref{eq:multi-head-attn}. We summarise these equations by:
\begin{equation}
    \label{eq:attention}
    \mcZ = \textrm{Attn}(\mcQ=\mcX, \mcV=\mcX) \equiv \textrm{SA}(\mcX).
\end{equation}

\subsection{Decoders}
\label{app:taxonomy-decoders}
For every decoder, we use when appropriate the same \gls{mlp}, composed of two layers: one from $\dhid$ to $\dhid$, and another from $\dhid$ to M, where $\dhid$ is either 8, 32 and 64 depending on the dataset, and M is the number of marks.

We also define the following terms for every decoder: $t$, a query time, $\bmz_t\equiv\bmz_{|\mcH_t|}$, the representation for the latest event in $\mcH_t$, i.e. the past event closest in time to $t$, and $t_i$, the time of the previous event.
In addition, we define $\bmq_t$, the representation of the query time $t$. 
For the cumulative models, $\bmq_t = \textrm{ParametricTemporal}(t)$, and for the Monte-Carlo models, $\bmq_t = \textrm{Temporal}(t)$, following \Cref{eq:parametric-temporal-embedding} and \Cref{eq:temporal-embedding}, respectively.

\paragraph{Conditional Poisson (CP)}
The conditional Poisson decoder returns:
\begin{align}
    \lambda_m^*(t) &= \text{\gls{mlp}}(\bmz_t), & \\
    \Lambda_m^*(t) &= \text{\gls{mlp}}(\bmz_t) (t - t_i),
\end{align}
where the \gls{mlp} is the same in both equations.

\paragraph{RMTPP}
Given the same elements as for the conditional Poisson, the RMTPP returns:
\begin{align}
  \lambda_m^*(t)
  &=\exp\left[ \bmW^{(1)}\bmz_t + w^{(2)} (t - t_i) + \bmb^{(1)}\right]_m \\
  \Lambda_m^*(t) &= \frac{1}{w^{(2)}} \left[ \exp(\bmW^{(1)}\bmz_t + \bmb^{(1)}) - \right. \\ &\quad \left. \exp(\bmW^{(1)}\bmz_t + w^{(2)} (t - t_i) + \bmb^{(1)}) \right]_m
\end{align}
where
$\bmW^{(1)}\in\mbbR^{\dhid\times M}$,
$w^{(2)}\in\mbbR$, and
$\bmb^{(1)}\in\mbbR^{M}$,
are learnable parameters.

\paragraph{Log-normal mixture (LNM)}

Following \cite{shchurIntensityFreeLearningTemporal2020}, the log-normal mixture model returns:

\begin{align}
  \bar p^*(t)
  &=\sum_{k=1}^Kw_k\frac1{(t -t_i) \, \sigma_k\sqrt{2\pi}} \nonumber\\ &\qquad \exp\left[ -\frac{(\log (t-t_i)-\mu_k)^2}{2\sigma_k^2}\right],
\end{align}
where $\bar p^*_m(t)=
  \bar p^*(t)\, \rho_m(\mcH_t)$, and  $\bmw,\mbsigma,\mbmu\in\mbbR^K$ are mixture weights, distribution means and
standard deviations, and are outputs of the encoder. These parameters are defined by:
\begin{align}
    \bmw &= \textrm{softmax}(\bmW^{(1)}\bmz_t + \bmb^{(1)}), 
    & \\
    \mbsigma &= \exp(\bmW^{(2)}\bmz_t + \bmb^{(2)}),
    & \\
    \mbmu &= \bmW^{(3)}\bmz_t + \bmb^{(3)}.
\end{align}

\paragraph{MLP Cumulative (MLP-CM)}
The MLP-CM returns:
\begin{align}
    \lambda_m^*(t) &= \frac{\partial\Lambda_m^*(t)}{\partial t}, & \\
    \Lambda_m^*(t) &= \textrm{\gls{mlp}}([\bmq_t, \bmz_t]; \mbtheta_{\geq\epsilon}),
\end{align}
where the square brackets indicate a concatenation.

\paragraph{MLP Monte Carlo (MLP-MC)}
The MLP-MC returns:
\begin{align}
    \lambda_m^*(t) &= \textrm{\gls{mlp}}([\bmq_t, \bmz_t]), & \\
    \Lambda_m^*(t) &= \textrm{MC}\left[\lambda_m^*(t^\prime),t^\prime,t_i,t\right],
\end{align}
where \gls{mc} represents the estimation of the integral $\int_{t_i}^{t}\lambda_m^*(t^\prime)\,dt^\prime$ using a Monte-Carlo sampling method.

\paragraph{Attention Cumulative (Attn-CM)}
For the Attn-CM and Attn-MC, the attention block differs from the SA encoder: the queries $\bmW_q \bmq$ are defined from the query representations, and the keys $\bmW_k \bmz$ and values $\bmW_v \bmz$ are defined from the encoder representations.

The Attn-CM returns:
\begin{align}
    \lambda_m^*(t) &= \frac{\partial\Lambda_m^*(t)}{\partial t}, & \\
    \Lambda_m^*(t) &= \textrm{Attn}(\mcQ=\{\bmq_t\}, \mcV=\mcZ),
\end{align}
where Transformer refers to \Cref{eq:attention}, with $\bmx = \bmq$ as input. Moreover, the components of this Attention are modified following \Cref{app:monotonic}. In particular the \gls{relu} activation is replaced by a Gumbel.

\paragraph{Attention Monte Carlo (Attn-MC)}
The Attn-MC returns:
\begin{align}
    \lambda_m^*(t) &= \textrm{Attn}(\mcQ=\{\bmq_t\}, \mcV=\mcZ), & \\
    \Lambda_m^*(t) &= \textrm{MC}\left[\lambda_m^*(t^\prime),t^\prime,t_i,t\right],
\end{align}
where \gls{mc} represents the estimation of the integral $\int_{t_i}^{t}\lambda_m^*(t^\prime)\,dt^\prime$ using a Monte-Carlo sampling method.

\section{Additional results}
\label{app:additional-results}

\subsection{Hawkes datasets}
The parameters of our Hawkes datasets are:
\begin{align}
    \text{Independent}&: \nonumber\\
    \mu &= 
    \begin{bmatrix} 0.1 \\ 0.05 \end{bmatrix},&\nonumber\\
    \alpha &= 
    \begin{bmatrix} 
    0.2 & 0.0 \\ 0.0 & 0.4 
    \end{bmatrix},&\nonumber\\
    \beta &=
    \begin{bmatrix} 
    1.0 & 1.0 \\ 1.0 & 1.0
    \end{bmatrix},\nonumber\\
    \text{Dependent}&: \\
    \mu &= 
    \begin{bmatrix} 0.1 \\ 0.05 \end{bmatrix},&\nonumber\\
    \alpha &= 
    \begin{bmatrix} 
    0.2 & 0.1 \\ 0.2 & 0.3 
    \end{bmatrix},&\nonumber\\
    \beta &=
    \begin{bmatrix} 
    1.0 & 1.0 \\ 1.0 & 2.0
    \end{bmatrix}.\nonumber
\end{align}

\subsection{Hawkes results}
A useful property of these datasets is that we can visually compare the modelled intensities with the intensity of the underlying generating process.
On each datasets, Hawkes (independent) and Hawkes (dependent), all models perform similarly, with the exception of the conditional Poisson. 
We present results for the SA-MLP-MC on \Cref{fig:hawkes-intensity}.

\begin{figure*}[!htb]
\small
\floatconts
  {fig:hawkes-intensity}
  {\caption{Intensity functions on the Hawkes datasets. The orange line represents the true intensity of the sequence, while the blue line represents the intensity modelled by the SA-MLP-MC.}}
  {%
    \subfigure[SA-MLP-MC on Hawkes (dependent)]{\label{fig:intensity-hawkes-1}%
      \includegraphics[width=0.5\linewidth]{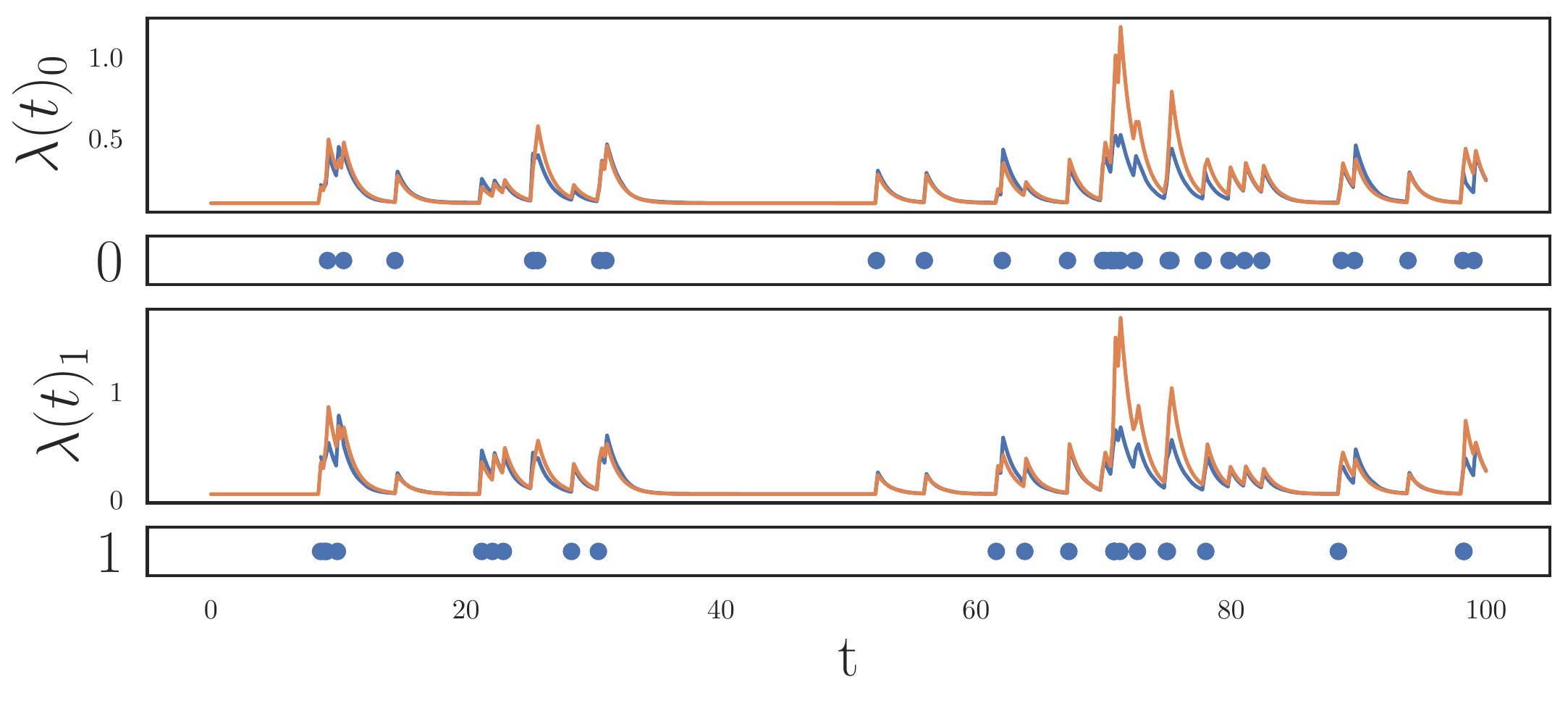}}%
    \subfigure[SA-MLP-MC on Hawkes (independent)]{\label{fig:intensity-hawkes-2}%
      \includegraphics[width=0.5\linewidth]{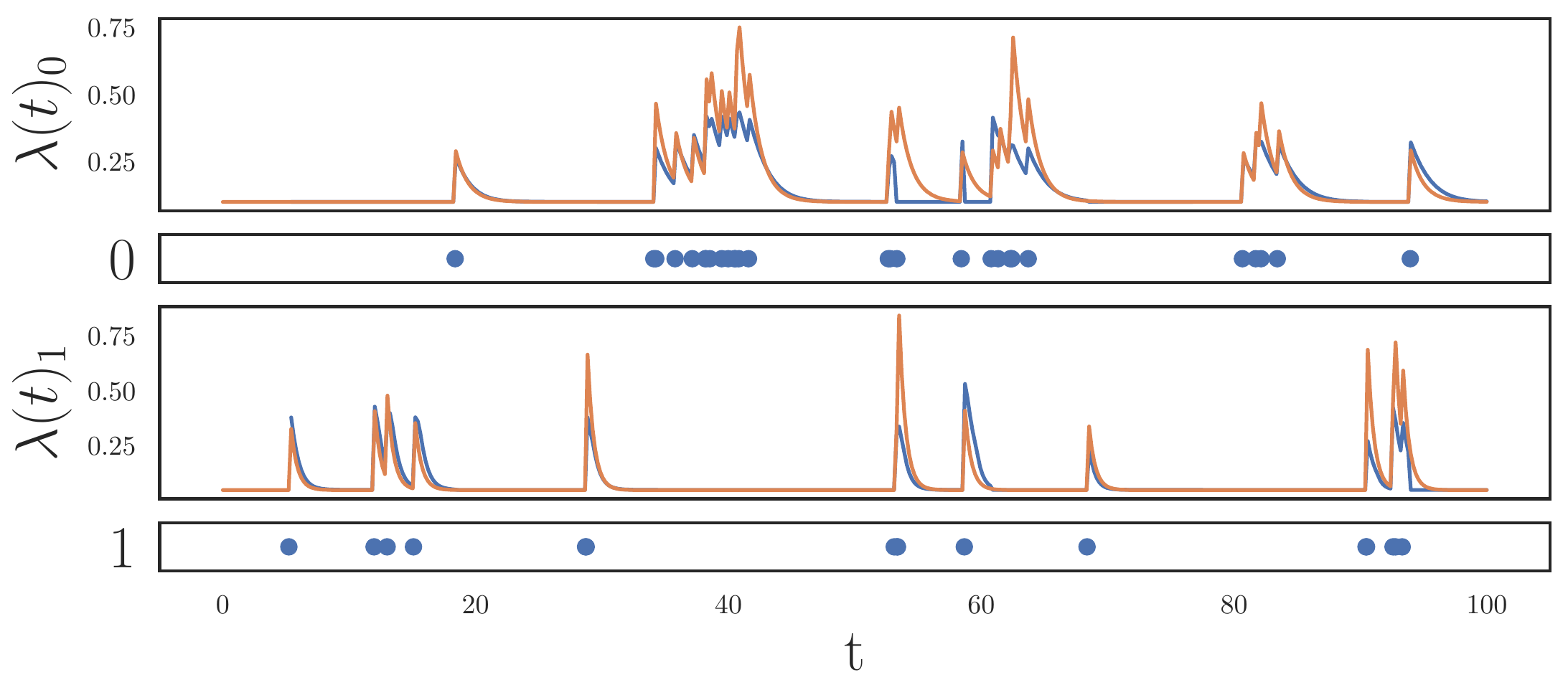}}
  }
\end{figure*}

\subsection{Retweets results}
On the Retweets dataset, while the conditional Poisson model significantly outperforms \gls{tpp} models in terms of ROC-AUC, the opposite is true in terms of NLL/time. This is probably due to the fact that \glspl{tpp} can model the intensity decay that occurs when a tweet is not retweeted over a certain period of time. 
We compare on \Cref{fig:retweets-intensity} the intensity functions of a SA-CP and and a SA-MLP-MC.

\begin{figure*}[!htb]
\small
\floatconts
  {fig:retweets-intensity}
  {\caption{Intensity functions of a SA-CP and a SA-MLP-MC on the Retweets datasets. The MLP-MC is able to model the time decay of the intensity, contrary to the CP.}}
  {%
    \subfigure[SA-CP on Retweets data]{\label{fig:intensity-retweets-cp}%
      \includegraphics[width=0.5\linewidth]{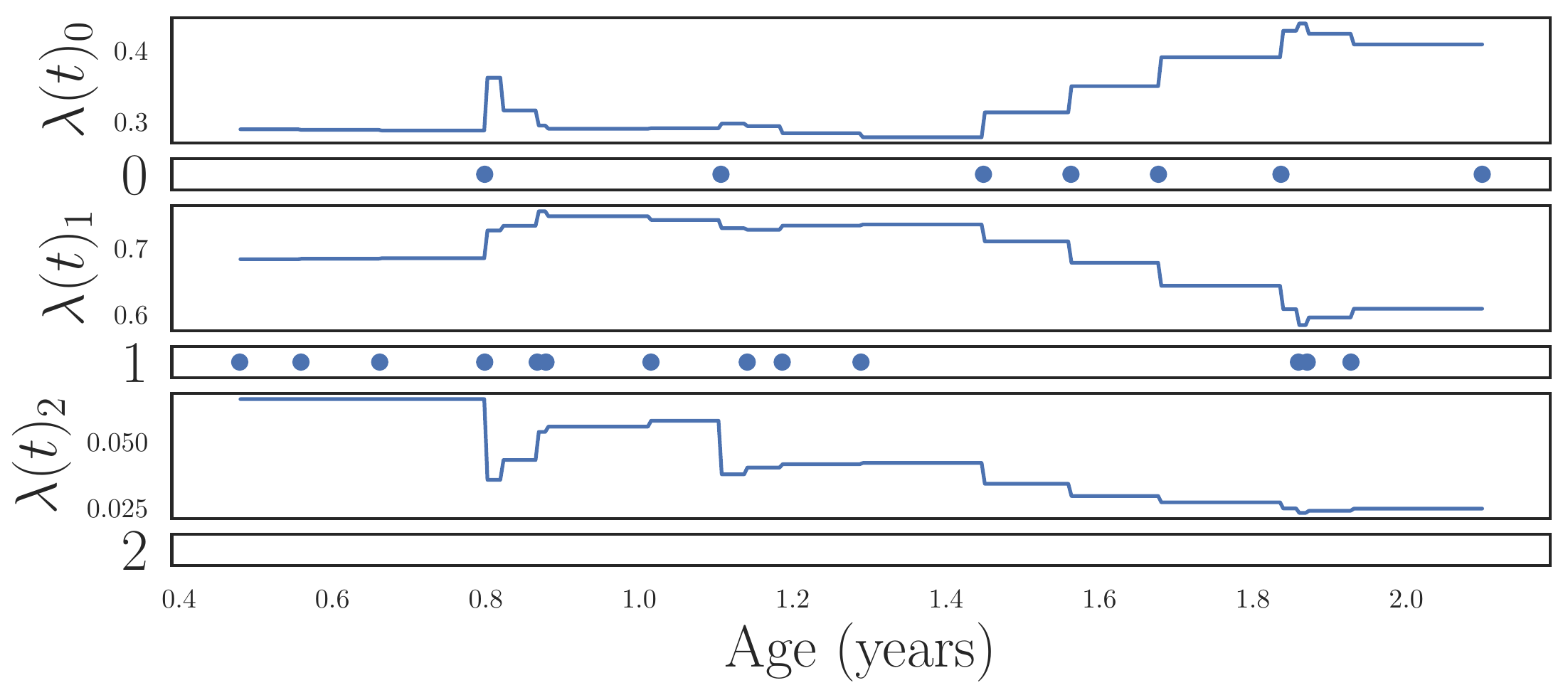}}%
    \subfigure[SA-MLP-MC on Retweets data]{\label{fig:intensity-retweets-mlp-mc}%
      \includegraphics[width=0.5\linewidth]{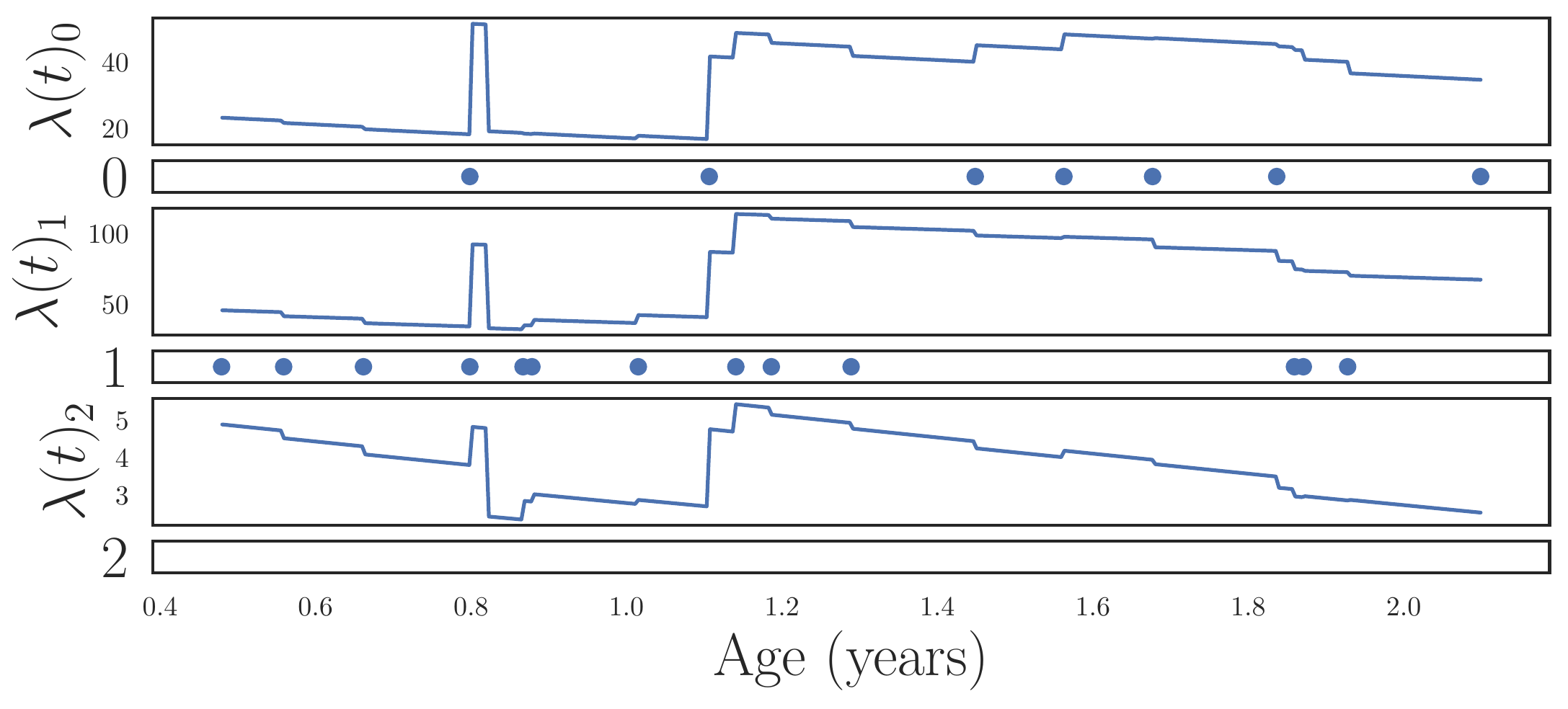}}
  }
\end{figure*}

\subsection{Attention coefficients}
\begin{figure*}[!htb]
\small
\centering
\caption{
Encoder attention coefficients for an \gls{ehr} extracted from the Synthea (Full) dataset.
Line thickness corresponds to attention strength, with values below 0.5 are omitted for clarity.
As stressed in \citet{serrano-smith-2019-attention}, ``while attention is by no means a fail-safe indicator", it is still able to "noisily predicts input components' overall importance to a model''.}
\includegraphics{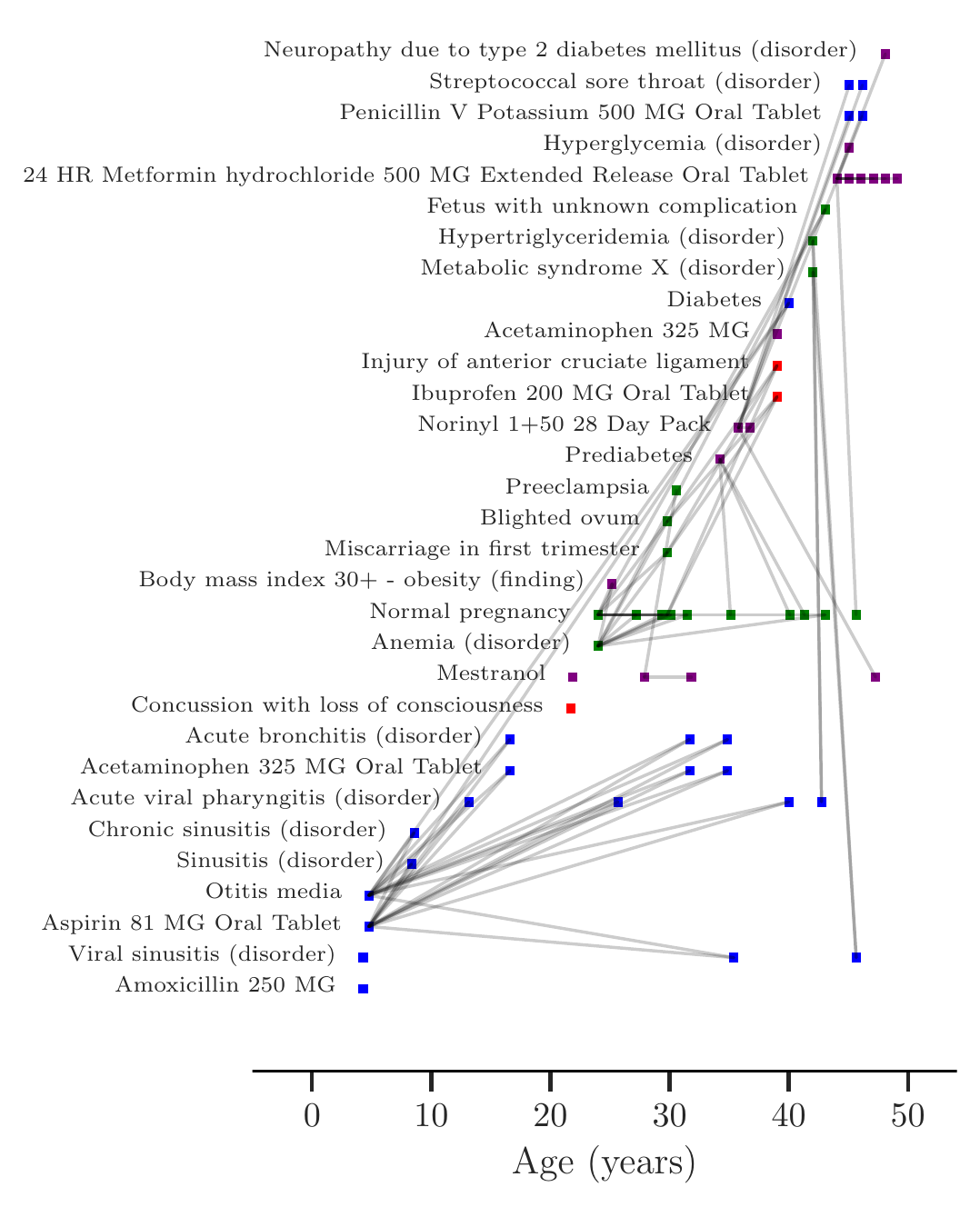}
\label{fig:synthea-attention}
\end{figure*}

\Cref{fig:synthea-attention} shows an example of attention coefficients for a sequence of Synthea (Full).
\section{Transformer architecture}
\paragraph{Attention block}
The key component of the Transformer is attention. This computes contextualised representations
$
\bmq_i^\prime
=\textrm{Attention}(\bmq_i,\{\bmk_j\},\{\bmv_j\})
\in\mbbR^{d_{\textrm{Model}}}
$
of queries
$\bmq_i\in\mbbR^{d_\textrm{Model}}$
from linear combinations of values
$\bmv_j\in\mbbR^{d_\textrm{Model}}$
whose magnitude of contribution is governed by keys
$\bmk_j\in\mbbR^{d_\textrm{Model}}$
\begin{align}
  \textrm{Attention}(\bmq_i,\{\bmk_j\},\{\bmv_j\})
  &=\sum_j \alpha_{i,j}\,\bmv_j,
  & \nonumber\\
  \alpha_{i,j}
  &=g\left(E_{i,j}\right),\label{attn}
  & \\
  E_{i,j}&=\frac{\bmq^T_i\bmk_j}{\sqrt{d_k}}\nonumber
\end{align}
where $\alpha_{i,j}$ are the attention coefficients, $E_{i,j}$ are the attention logits, and $g$ is an activation function that is usually taken to be the softmax
\begin{equation}
  \softmax_j\left( E_{i,j} \right)=
  \frac{\exp(E_{i,j})}{\sum_k\exp(E_{i,k})}.
\label{attn_activation}
\end{equation}

\paragraph{Masking}
The masking in our encoder is such that events only attend to themselves and events that happened strictly before them in time. The masking in the decoder is such that the query only attends to events occurring strictly before the query time.

\paragraph{Temporal Encoding}
In the original transformer \citep{vaswaniAttentionAllYou2017}, absolute positional information is encoded into $\bmq_i$, $\bmk_j$ and $\bmv_j$. This was achieved by producing positional embeddings where the embeddings of relative positions were linearly related $\bmx(i) = \bmR(i-j) \, \bmx(j)$ for some rotation matrix $\bmR(i-j)$.
Our temporal embedding method generalises this to the continuous domain
\begin{multline}
  \label{eq:temporal-embedding}
  \textrm{Temporal}(t)
  = \\
    \bigoplus_{k=0}^{d_{\textrm{Model}}/2-1}\sin(\alpha_k t)\oplus \cos(\alpha_k t)\in\mbbR^{d_\textrm{Model}},
\end{multline}
where $\alpha_k$ is proportional to  $10000^{-2k/d_{\textrm{Model}}}$, and $\beta$ is a temporal rescaling parameter and plays the role of setting the shortest time scale the model is sensitive to. In practice we estimate $\hat\beta=\mbbE[(w_+-w_-)/N]$ from the training set, so that a \gls{tpp} with smaller average gaps between events is modelled at a higher granularity by the encoding. We experimented with having $\beta$ a learnable parameter of the model, however, this made the model extremely difficult to optimise.
Note that $\hat\beta=1$ for a language model.
In addition, $\beta$ does not change the relative frequency of the rotational subspaces in the temporal embedding from the form in \citet{vaswaniAttentionAllYou2017}.
The encoding with temporal information of an event at $t_i$ with labels $\mcM_i$ is then
\begin{multline}
  \label{eq:temporal-encoding}
  \bmx_i
  = \textrm{TemporalEncode}(t_i,\mcM_i) = \\ \bmv_i(\mcM_i)\sqrt{d_{\textrm{Model}}} + \textrm{Temporal}(t_i)\in\mbbR^{d_{\textrm{Model}}}.
\end{multline}

\paragraph{Multi-head attention}
Multi-head attention produces $\bmq_i^\prime$ using $H$ parallel attention layers (heads) in order to
jointly attend to information from different subspaces at different positions
\begin{multline}
\label{eq:multi-head-attn}
\textrm{MultiHead}(\bmq_i,\{\bmk_j\},\{\bmv_j\})
= \\ \bmW^{(o)}\Bigg[\bigoplus_{h=1}^H
  \textrm{Attn}\Big(\bmW_h^{(q)}\bmq_i,\{\bmW_h^{(k)}\bmk_j\},\\
  \{\bmW_h^{(v)}\bmv_j\}
  \Big)\Bigg],
\end{multline}
where
$\bmW_h^{(q)},\bmW_h^{(k)}\in\mbbR^{d_k\times d_\textrm{Model}}$,
$\bmW_h^{(v)}\in\mbbR^{d_v\times d_\textrm{Model}}$,
and
$\bmW^{(o)}\in\mbbR^{d_\textrm{Model}\times h\,d_v}$
are learnable projections.

\section{Broader impact statement}
\label{sec:impact}
\subsection{Overview}
In this paper, we demonstrate that Temporal Point Processes perform favourably in modelling Electronic Health Records. We embarked on this project aware that as high impact systems, Clinical Decision Support tools must ultimately provide explanations \citep{holzingerWhatWeNeed2017} to ensure clinician adoption \citep{millerExplanationArtificialIntelligence2019} and provide accountability \citep{mittelstadtEthicsAlgorithmsMapping2016} whilst being mindful of the call for researchers to build interpretable ML models \citep{rudinStopExplainingBlack2019}. Our results demonstrate that our attention-based model performs favourably against less interpretable models. As attention could carry medically interpretable information our innovation contributes towards the development of medically useful AI tools.

Aware of both the need for transparency in the research community \citep{pineauImprovingReproducibilityMachine2020} and for sensitive handling of health data \citep{kalkmanResponsibleDataSharing2019}, we opted to use open source synthetic EHRs. We provide all trained models, benchmarked datasets and a high-quality deterministic codebase to allow others to easily implement and benchmark their own models. We also make the important contribution of highlighting existing benchmark datasets as inappropriate; continuing to develop temporal models using these datasets may slow development of useful technologies or cause poor outputs if applied to EHR data.

Turning to the potential impact of our technology, we begin by briefly discussing the benefits temporal EHR models could bring to healthcare before, with no less objectivity, analysing harms which could occur, raising questions for further reflection. 

\subsection{The potential benefits of EHR modelling}
Healthcare systems today are under intense pressure. Chronic disease prevalence is rising \citep{raghupathiEmpiricalStudyChronic2018}, costs of care are increasing, resources are constrained and outcomes are worsening \citep{topolHighperformanceMedicineConvergence2019}. 23\% of UK deaths in 2017 were classed as avoidable \citep{officefornationalstatisticsAvoidableMortalityUK2017}, clinician diagnostic error is estimated at 10-15\% \citep{graberIncidenceDiagnosticError2013} and EHR data suffers from poor coding and incompleteness which affects downstream tasks \citep{jetleyElectronicHealthRecords2019}. Our research contributes techniques which could be used to:
\begin{itemize}
    \item More accurately impute missing EHR data.
    \item Identify likely misdiagnoses. 
    \item Predict future health outcomes.
    \item Identify higher risk patients for successful health interventions.
    \item Design optimal care pathways.
    \item Optimise resource management.
    \item Identify previously undiscovered links between health conditions.
\end{itemize}
Clearly, implemented well, temporal EHR modelling could provide immense benefit to our society. These benefits outweigh possible harms on the provision that they are well mitigated. Hence, we devote the rest of the discussion to evaluating potential harms.

\subsection{Potential Harms of EHR Modelling}
There are a number of ways in which the temporal modelling of EHRs could create societal harm. To examine these risks, we evaluate them by topic, framed as a series of case studies.

\subsubsection{Poor Health Outcomes Through Data Bias}
Bias can mean many things when discussing the application of AI to healthcare. As such, we would encourage researchers to use \citet{sureshFrameworkUnderstandingUnintended2020}'s framework to better identify and address the specific challenges that arise when modelling health data. All are relevant to our technology, but we will specifically focus on issues arising from historical, representation and aggregation biases. Importantly, we anticipate these biases would be more easily identified due to the interpretability of our attention-based technique. 

\paragraph{Scenario} Temporal EHR modelling is deployed to predict future health having been trained on historical health records.


\paragraph{Societal harm} 
Societal groups who have historically been discriminated against continue to receive below standard healthcare.

\paragraph{Examples of historical biases}
Historically, the majority of medical trials have been conducted on white males \citep{morleyDebateEthicsAI2019} providing greater health information on this population group. We also know that clinicians are fallible to human biases. For example: women are less likely than men to receive optimal care despite being more likely to present with hypertension and heart failure \citep{lishanshanSexRaceEthnicity2016}; They are often not diagnosed with diseases due to human bias \citep{sureshFrameworkUnderstandingUnintended2020} and are less likely than men to be given painkillers when they report they are in pain \citep{calderoneInfluenceGenderFrequency1990}; African Americans more likely to be misdiagnosed with schizophrenia than white patients \citep{garaNaturalisticStudyRacial2018}.

\paragraph{Societal harm}
Societal groups who have historically been discriminated against continue to receive below standard healthcare.

\paragraph{Examples of historical biases} 
Historically, the majority of medical trials have been conducted on white males \citep{morleyDebateEthicsAI2019} providing greater health information on this population group. We also know that clinicians are fallible to human biases. For example: women are less likely than men to receive optimal care despite being more likely to present with hypertension and heart failure \citep{lishanshanSexRaceEthnicity2016}; They are often not diagnosed with diseases due to human bias \citep{sureshFrameworkUnderstandingUnintended2020} and are less likely than men to be given painkillers when they report they are in pain \citep{calderoneInfluenceGenderFrequency1990}; African Americans more likely to be misdiagnosed with schizophrenia than white patients \citep{garaNaturalisticStudyRacial2018}.

\begin{center}
    * * *
\end{center}
\paragraph{Scenario} Temporal EHR modelling is deployed to predict future health and develop care plans. 
One model is used for a diverse population group.

\paragraph{Societal harm} Demographics' varying symptom patterns and care plan needs are not accounted for, resulting in below optimal health outcomes for the majority.
\paragraph{Examples of aggregation biases} It has been shown that the risk factors for carotid artery disease differ by ethnicity \citep{gijsbertsRaceEthnicDifferences2015} yet prevention plans are developed on data from almost exclusively a white population. A common measurement for diabetes widely used for diagnosis and monitoring differs in value across ethnicities and genders \citep{hermanRacialEthnicDifferences2012}.
\begin{center}
    * * *
\end{center}
\paragraph{Scenario} Researchers select proxies to represent metrics within their model without considering broader socioeconomic factors.

\paragraph{Societal harm} Societal groups are subject to allocation biases where they do not receive the same standard of healthcare compared to advantaged groups.

\paragraph{Example of poor proxy selection} \citep{obermeyerDissectingRacialBias2019} analysed a model in current commercial use which identifies high risk patients for care interventions used medical costs as a proxy for illness. It was found that black people were much sicker than white people with the same risk score as historical and socioeconomic factors mean that black people do not receive parity in health care and treatments. 

\begin{center}
    * * *
\end{center}
\paragraph{Scenario} Biased models are deployed as clinician-in-the-loop with the intention of using a human to protect against model biases.

\paragraph{Societal harm} The clinician cannot redress model biases, causing the above mentioned harms to persist.
\paragraph{Example of human-in-the-loop failure} Humans are subject to confirmation bias \citep{greenDisparateInteractionsAlgorithmintheLoop2019} and \citep{obermeyerDissectingRacialBias2019} found that whilst clinicians can redress some disparity in care allocation, they do not adjust the balance to that of a fair algorithm.

\begin{center}
    * * *
\end{center}
Recommendations: 
\begin{itemize}
    \item Researchers must employ caution when using proxies such as ``diagnosed with condition X’’ equals ``has condition X’’.
    \item Models must be evaluated against the intended recipient populations. 
    \item We call on all institutions deploying temporal EHR models to assess fair performance in the target population, and, on policy makers to provide benchmarked patient datasets on which fair and safe performance of models must be demonstrated before deployment.
\end{itemize}

Questions for reflection:
\begin{itemize}
    \item How do we best formulate the problems temporal EHR modelling is trying to solve to prevent discriminatory harms?
    \item If temporal EHR modelling was deployed, what techniques would be used to protect society from the reinforcement of structural inequalities and structural injustices \citep{jugovStructuralInjusticeEpistemic2019}?
    \item How do you construct adequate benchmarking datasets to test the suitability of a temporal EHR model for a given population?
    \item What evidence is needed to demonstrate the clinical effectiveness of a temporal model \citep{greavesWhatAppropriateLevel2018}?
\end{itemize}

\subsubsection{Poor care provision through lack of contextual awareness}

\paragraph{Scenario} A temporal model is trained in a perfectly unbiased way and is deployed for use.
\paragraph{Societal harm} As the model is not aware of emergent health outcomes, information provided is out of sync with modern medical knowledge. Potentially resulting in sub-optimal care plan recommendations and predictions.
\paragraph{Example of lack of contextual awareness} Consider training on a cohort of 100-year-old patients. They lived through the second world war, ate rations, never took a contraceptive pill and were retiring as computers became commonplace. Individuals approaching old age in the future will likely have very different health trajectories. By their nature, modelling of EHR data will be context blind \citep{caruanaIntelligibleModelsHealthCare2015}. For example, if a new drug is found to prevent the development of Type 2 diabetes, model predictions will be out of sync with patient outcomes, it would view the prescription of this better drug as an error.

\begin{center}
    * * *
\end{center}
Questions for reflection:
\begin{itemize}
    \item How would we develop temporal EHR models to correct dynamically for non-stationary processes (changing medical knowledge)?
    \item How do we introduce contextual awareness to temporal EHR models?
\end{itemize}

\subsubsection{Restrictions in health access}

\paragraph{Scenario} A perfect model is deployed to be used by health insurers to identify high cost patients. Those who can be assisted by interventions are helped. The model identifies a set of individuals who are unavoidably costly, due to health issues beyond their control. 

\paragraph{Societal harm} This may cause a shift in the economic model of the society in which the EHR model is deployed \citep{balasubramanianInsurance2030Impact2018}.

\paragraph{Examples of economic impact} Temporal models will make increasingly accurate predictions about how much each individual's healthcare will cost over a lifetime. In countries where healthcare is not universal insurers and providers will adjust their actuarial models, moving away from pooled risk healthcare costs. Health insurers increase premiums to an unfeasibly high amount for costly individuals, preventing their access to healthcare and pricing models change for those who are able to access care. Governments are forced to determine whether to use social security to provide universal healthcare. If it is not the divide between those who can access healthcare and those who can’t potentially widens with ripple effects across the broader economy.  

\begin{center}
    * * *
\end{center}
Questions for reflection:
\begin{itemize}
    \item How do we as a society determine how to carry the cost of healthcare provision?
    \item What level of risk granularity should health insurance companies be privy to?
\end{itemize}

\subsubsection{Changes to the Doctor / Patient relationship and agency}

\paragraph{Scenario} Temporal EHR modelling causes a paradigm shift towards solely data-centred medicine. Reducing a patient solely to their datapoints. Theoretically, the model can predict a patient’s death.

\paragraph{Societal harm}Agency is removed from both the patient and the doctor. Knowledge of death raises existential questions and changes individuals’ behaviours.
\paragraph{Example of solely data-centred medicine} The doctor is left to evaluate symptoms solely on datapoints \citep{kleinpeterFourEthicalIssues2017}. A patient’s ability to fully express what it means to inhabit their body and feel illness (or not) is removed. They may feel unable to decline or deviate from recommended treatment plans, impacting the patient’s perception of bodily autonomy. 

\begin{center}
    * * *
\end{center}
\paragraph{Scenario} Temporal EHR models are deployed in  consumer health and wellness applications with the current narrative of empowering an individual to take control of their health \citep{morleyEnablingDigitalHealth2019}. 

\paragraph{Societal harm} The burden of disease prevention is shifted to the individual who has proportionally little control over macro health issues, tantamount to victim blaming \citep{morleyEnablingDigitalHealth2019}.
\paragraph{Example of empowerment ineffectiveness} Many factors of health are not solvable with applications. For example, those on low incomes can’t afford healthy foods and other socioeconomic factors prevent individuals from accessing healthcare \citep{defreitasInclusivePublicParticipation2015}. 

\begin{center}
    * * *
\end{center}
Recommendations:
\begin{itemize}
    \item Policy makers should contemplate how far into the future various institutions should be permitted to predict peoples’ health.
    \item We encourage future researchers to view consumer applications that use temporal EHR modelling as digital companions \citep{morleyLimitsEmpowermentHow2019}. 
\end{itemize}
Questions for reflection:
\begin{itemize}
    \item How many years in advance should predictive health technologies be able to predict?
    \item How much clinical decision-making should we be delegating to AI-Health solutions \citep{nucciShouldWeBe2019}?
    \item How can we develop temporal EHR models to account for the socioeconomic and historical factors which contribute to poor health outcomes?
    \item How do we incorporate into temporal EHR modelling an ethical focus on the end user, and their expectations, demands, needs, and rights \citep{morleyEthicallyMindfulApproach2020}?
\end{itemize}

\subsection{Wider reaching applications}

The temporal modelling of irregular data could be applied to many domains. For example, education and personalised learning, crime and predictive policing, social media interactions and more. Each of these applications raises questions around the societal impact of improved temporal modelling. We hope that the above analysis provides stimulation when developing its beneficial usage.


\appendix
\end{document}